\documentclass{article}
\PassOptionsToPackage{sort&compress}{natbib}
\usepackage{iclr2027_conference,times}
\setcitestyle{numbers,square,citesep={,}}

\usepackage{amsmath,amsfonts,bm}

\def\eqref#1{equation~\ref{#1}}

\def\1{\bm{1}}

\DeclareMathAlphabet{\mathsfit}{\encodingdefault}{\sfdefault}{m}{sl}
\SetMathAlphabet{\mathsfit}{bold}{\encodingdefault}{\sfdefault}{bx}{n}

\usepackage[utf8]{inputenc}
\usepackage[T1]{fontenc}
\usepackage{amsmath,amssymb,amsthm}
\usepackage{graphicx}
\usepackage{booktabs}
\usepackage{xcolor}
\usepackage{listings}
\usepackage{hyperref}
\usepackage{url}
\usepackage[capitalize,noabbrev]{cleveref}
\usepackage{tikz}
\usepackage{placeins}
\usepackage[inline,shortlabels]{enumitem}
\usetikzlibrary{arrows.meta,positioning}
\usepackage{sansmath}

\definecolor{wspink}{HTML}{0B0B0B}
\definecolor{wspinktwo}{HTML}{52514E}
\definecolor{wspgrid}{HTML}{E4E3DF}
\definecolor{wspdata}{HTML}{A3A29D}
\definecolor{wspbefore}{HTML}{2A78D6}
\definecolor{wspafter}{HTML}{EB6834}
\newcommand{\figurefont}{\fontsize{8}{9.5}\selectfont}

\graphicspath{{figures/}}

\newcommand{\equal}{\textsuperscript{\rm *}}
\newcommand{\equaltext}{\equal\ Equal contribution}

\newcommand{\ufr}{\textsuperscript{\rm 1}}
\newcommand{\ufrc}{\textsuperscript{\rm 1,}}
\newcommand{\ufrtext}{\ufr\ University of Freiburg}

\newcommand{\prior}{\textsuperscript{\rm 2}}
\newcommand{\priorc}{\textsuperscript{\rm 2,}}
\newcommand{\priortext}{\prior\ Prior Labs}

\newcommand{\ellis}{\textsuperscript{\rm 3}}
\newcommand{\ellisc}{\textsuperscript{\rm 3,}}
\newcommand{\ellistext}{\ellis\ ELLIS Institute Tübingen}

\title{Weights \& Caches: LLMs Can Program Tabular Foundation Models}

\author{Shi Bin Hoo\ufrc\equal \quad Magnus Bühler\ufrc\equal \quad Omar Swelam\ufrc\equal\\
\textbf{Frank Hutter}\priorc\ellisc\ufr \quad \textbf{Lennart Purucker}\priorc\ufr\\
\ufrtext, \priortext, \ellistext \\
\texttt{\{hoos,buehlema,swelamo\}@cs.uni-freiburg.de}
}

\iclrfinalcopy

\begin{document}

\maketitle
{\renewcommand{\thefootnote}{}\footnotetext{\equaltext}}

\begin{abstract}

Neural networks are typically programmed through optimization, rather than by explicitly specifying their computations. Recent work on weight-space programming suggests that large language models (LLMs) can instead reason about and manipulate neural-network weights directly. In this work, we show that LLMs can directly program statistical computations in tabular foundation models (TFMs), whose parameters enable inference across different datasets. We investigate three complementary settings using GPT-6 Astra, without gradient-based optimization.
First, Astra directly programs nanoTabPFN from zero-initialized weights, achieving a mean ROC AUC of $0.849$ across 20 TabArena datasets, compared to $0.876$ for a pretrained baseline.
Second, Astra diagnoses and repairs a signed-product regression failure in pretrained TabPFN-3.5, restoring near-perfect predictions on held-out synthetic datasets while preserving performance on TabArena-Lite, with only a $0.08\%$ average relative increase in error.
Third, it writes KV-cache-editing programs that steer TabPFN-3.5's predictions on unseen synthetic datasets, demonstrating transformations such as reversing linear trends and converting sine waves into approximate square waves, without modifying model weights.
Together, our findings suggest that LLM-driven programming could complement gradient-based optimization across TFMs' development and usage, while offering a new way to investigate and intervene in their internal computations.

\end{abstract}

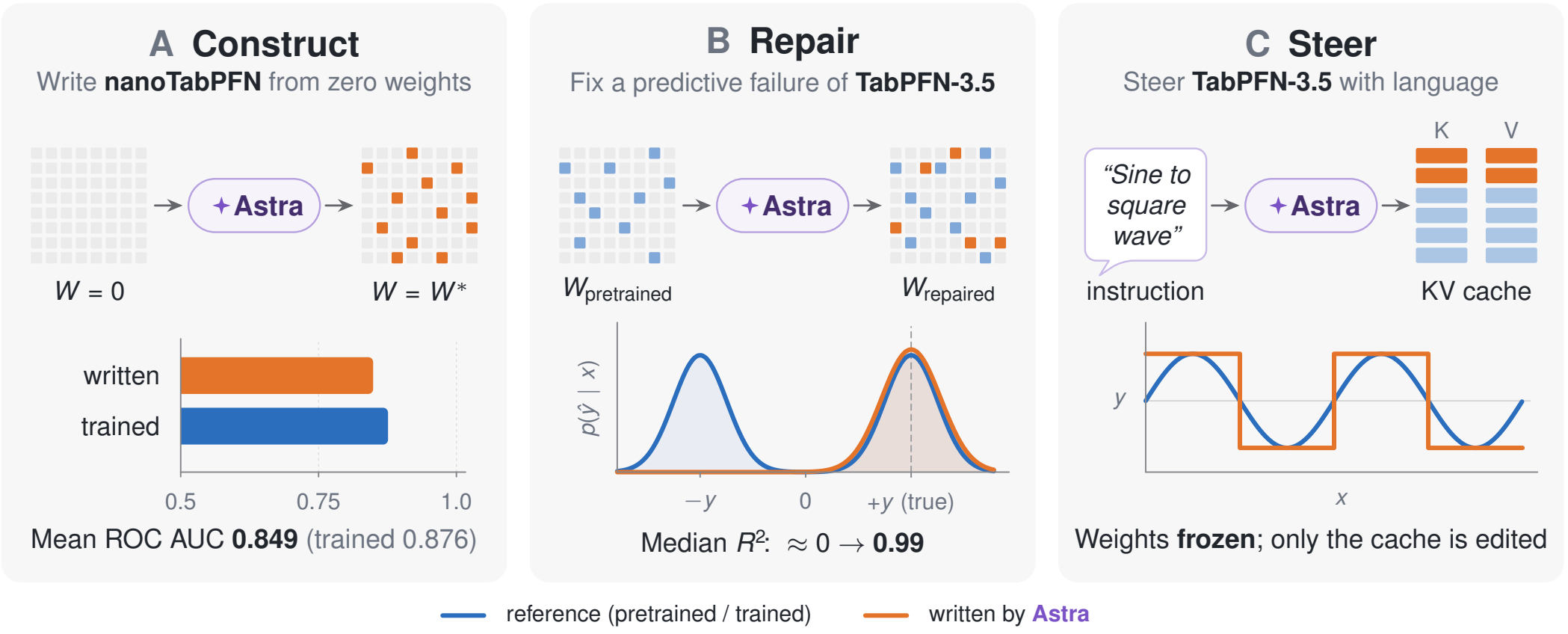
\begin{figure}[h]
    \centering
    \resizebox{\linewidth}{!}{%
        \begin{tikzpicture}[x=1cm, y=1cm, line cap=round, line join=round,
  title/.style    ={font=\sffamily\sansmath\bfseries\fontsize{11}{13}\selectfont, text=ink},
  sub/.style      ={font=\sffamily\sansmath\fontsize{8}{9.5}\selectfont, text=muted},
  lab/.style      ={font=\sffamily\sansmath\fontsize{8}{9.5}\selectfont, text=ink},
  tick/.style     ={font=\sffamily\sansmath\fontsize{7}{8}\selectfont, text=muted},
  result/.style   ={font=\sffamily\sansmath\fontsize{8.2}{10}\selectfont, text=ink},
  astranode/.style={draw=astra!55, fill=astra!7, rounded corners=2.9mm, line width=.55pt,
                    inner xsep=5pt, inner ysep=4pt, minimum height=.58cm,
                    font=\sffamily\bfseries\fontsize{8.5}{10}\selectfont, text=astra!55!black},
  flow/.style     ={-{Stealth[length=1.7mm,width=1.4mm]}, line width=.65pt, ink!70},
  axis/.style     ={line width=.45pt, ink!55},
  guide/.style    ={line width=.3pt, black!14, dash pattern=on .6pt off 1.2pt},
  curve/.style    ={line width=1.4pt},
]
\definecolor{orig}{RGB}{44,110,190}
\definecolor{edit}{RGB}{228,116,42}
\definecolor{astra}{RGB}{122,82,198}
\definecolor{ink}{RGB}{33,37,44}
\definecolor{muted}{RGB}{108,115,128}

\def\pw{5.5}
\def\ph{6.3}
\def\rowy{4.1}
\def\labely{3.17}
\def\bx{.95}
\def\ax{4.55}
\def\cs{.1625}
\def\gap{.02}

\def\wspgrid#1#2#3{%
  \begin{scope}[shift={#1}, shift={(-4*\cs,-4*\cs)}]
    \foreach \x in {0,...,7}{\foreach \y in {0,...,7}{
      \fill[black!8, rounded corners=.4pt]
        (\x*\cs+\gap,\y*\cs+\gap) rectangle ++(\cs-2*\gap,\cs-2*\gap);}}
    \def\tmp{#2}\ifx\tmp\empty\else
      \foreach \x/\y in {#2}{\fill[orig!60, rounded corners=.4pt]
        (\x*\cs+\gap,\y*\cs+\gap) rectangle ++(\cs-2*\gap,\cs-2*\gap);}\fi
    \def\tmp{#3}\ifx\tmp\empty\else
      \foreach \x/\y in {#3}{\fill[edit, rounded corners=.4pt]
        (\x*\cs+\gap,\y*\cs+\gap) rectangle ++(\cs-2*\gap,\cs-2*\gap);}\fi
  \end{scope}}

\def\spark#1{\begin{scope}[shift={#1}]
  \fill[fill=astra, rounded corners=.25pt] (0,.1) -- (.023,.023) -- (.1,0) -- (.023,-.023)
    -- (0,-.1) -- (-.023,-.023) -- (-.1,0) -- (-.023,.023) -- cycle;\end{scope}}

\def\wsppanel#1#2#3#4{%
  \fill[black!3, rounded corners=6pt] (#1,0) rectangle ++(\pw,\ph);
  \node[title] at (#1+\pw/2, 5.86) {\textcolor{muted}{#2}\enspace #3};
  \node[sub]   at (#1+\pw/2, 5.42) {#4};
  \node[astranode] (A) at (#1+\pw/2, \rowy) {\hspace{9pt}Astra};
  \spark{([xshift=10.5pt]A.west)}
  \draw[flow] (#1+\bx+.72,\rowy) -- ([xshift=-1.5pt]A.west);
  \draw[flow] ([xshift=1.5pt]A.east) -- (#1+\ax-.72,\rowy);}

\begin{scope}
  \wsppanel{0}{A}{Construct}{Write \textcolor{ink}{\textbf{nanoTabPFN}} from zero weights}
  \wspgrid{(\bx,\rowy)}{}{}
  \wspgrid{(\ax,\rowy)}{}{0/6,1/2,2/4,3/7,3/1,4/5,5/0,5/3,6/6,7/2,7/4,2/0}
  \node[lab] at (\bx,\labely) {$W = 0$};
  \node[lab] at (\ax,\labely) {$W = W^{*}$};
  \begin{scope}[shift={(1.95,1.2)}]
    \def\sx{6.0}
    \foreach \v in {1.5, 3.0}{\draw[guide] (\v,0) -- (\v,1.45);}
    \fill[edit, rounded corners=1.5pt] (-.2,.85) rectangle ({(0.849-.5)*\sx},1.25);
    \fill[orig, rounded corners=1.5pt] (-.2,.30) rectangle ({(0.876-.5)*\sx},.70);
    \fill[black!3] (-.25,.2) rectangle (0,1.35);
    \node[lab, anchor=east] at (-.1,1.05) {written};
    \node[lab, anchor=east] at (-.1,.50) {trained};
    \draw[axis] (0,0) -- (3.1,0);
    \draw[axis] (0,0) -- (0,1.45);
    \foreach \v/\t in {0/0.5, 1.5/0.75, 3.0/1.0}{
      \draw[axis] (\v,0) -- (\v,-.05);
      \node[tick, anchor=north] at (\v,-.1) {\t};}
  \end{scope}
  \node[result] at (\pw/2,.45) {Mean ROC AUC \textbf{0.849} \textcolor{muted}{(trained 0.876)}};
\end{scope}

\begin{scope}[shift={(5.75,0)}]
  \wsppanel{0}{B}{Repair}{Fix a predictive failure of \textcolor{ink}{\textbf{TabPFN-3.5}}}
  \wspgrid{(\bx,\rowy)}{0/6,1/1,2/3,3/6,4/2,5/4,6/0,7/5,1/4,6/7}{}
  \wspgrid{(\ax,\rowy)}{0/6,1/1,2/3,3/6,4/2,5/4,6/0,7/5,1/4,6/7}{2/6,4/7,5/1,0/2,7/1}
  \node[lab] at (\bx,\labely) {$W_{\mathrm{pretrained}}$};
  \node[lab] at (\ax,\labely) {$W_{\mathrm{repaired}}$};
  \begin{scope}[shift={(.95,1.2)}, x=4.1cm, y=1.55cm]
    \fill[orig, opacity=.07, smooth, samples=90, domain=0:1]
      plot (\x,{.82*exp(-((\x-.22)/.10)^2)+.82*exp(-((\x-.78)/.10)^2)}) -- (1,0) -- (0,0) -- cycle;
    \fill[edit, opacity=.13, smooth, samples=90, domain=0:1]
      plot (\x,{.86*exp(-((\x-.78)/.11)^2)}) -- (1,0) -- (0,0) -- cycle;
    \draw[ink!45, line width=.45pt, dash pattern=on 1.8pt off 1.4pt] (.78,0) -- (.78,1.02);
    \draw[orig, curve, smooth, samples=90, domain=0:1]
      plot (\x,{.82*exp(-((\x-.22)/.10)^2)+.82*exp(-((\x-.78)/.10)^2)});
    \draw[edit, curve, smooth, samples=90, domain=0:1]
      plot (\x,{.86*exp(-((\x-.78)/.11)^2)});
    \draw[axis] (0,0) -- (1.04,0);
    \draw[axis] (0,0) -- (0,1.05);
    \foreach \v/\t in {.22/$-y$, .5/$0$, .78/$+y$ (true)}{
      \draw[axis] (\v,0) -- (\v,-.035);
      \node[tick, anchor=north] at (\v,-.065) {\t};}
    \node[tick, rotate=90, anchor=south] at (-.015,.52) {$p(\hat y \mid x)$};
  \end{scope}
  \node[result] at (\pw/2,.45) {Median $R^2$\!:\; $\approx 0 \rightarrow$ \textbf{0.99}};
\end{scope}

\begin{scope}[shift={(11.5,0)}]
  \wsppanel{0}{C}{Steer}{Steer \textcolor{ink}{\textbf{TabPFN-3.5}} with language}
  \filldraw[fill=white, draw=astra!40, line width=.55pt, rounded corners=3.5pt]
    (\bx-.66,\rowy-.6) -- (\bx-.66,\rowy+.62) -- (\bx+.66,\rowy+.62) -- (\bx+.66,\rowy-.6)
    -- (\bx-.26,\rowy-.6) [sharp corners] -- (\bx-.5,\rowy-.76) -- (\bx-.4,\rowy-.6)
    [rounded corners=3.5pt] -- cycle;
  \node[align=center, font=\sffamily\itshape\fontsize{8}{9.5}\selectfont, text=ink]
    at (\bx,\rowy+.01) {``Sine to\\square\\wave''};
  \foreach \c/\xx in {K/3.89, V/4.65}{
    \foreach \r in {0,...,5}{
      \pgfmathsetmacro\col{\r>3 ? "edit" : "orig!40"}
      \fill[\col, rounded corners=.6pt] (\xx, \rowy-.65+\r*.2167+.025) rectangle ++(.56,.1667);}
    \node[tick] at (\xx+.28, \rowy+.82) {\c};}
  \node[lab] at (\bx,\labely) {instruction};
  \node[lab] at (\ax,\labely) {KV cache};
  \begin{scope}[shift={(.95,1.2)}, x=4.1cm, y=1.55cm]
    \draw[black!18, line width=.4pt] (0,.5) -- (1.02,.5);
    \draw[orig, curve, smooth, samples=120, domain=0:1]
      plot (\x,{.5+.33*sin(720*\x)});
    \draw[edit, curve]
      (0,.83) -- (.25,.83) -- (.25,.17) -- (.5,.17) -- (.5,.83)
      -- (.75,.83) -- (.75,.17) -- (1,.17);
    \draw[axis] (0,0) -- (1.04,0);
    \draw[axis] (0,0) -- (0,1.05);
    \node[tick, anchor=north] at (.52,-.065) {$x$};
    \node[tick, anchor=east] at (-.02,.5) {$y$};
  \end{scope}
  \node[result] at (\pw/2,.45) {Weights \textbf{frozen}; only the cache is edited};
\end{scope}

\begin{scope}[shift={(8.5,-.36)}]
  \draw[orig, curve] (-3.7,0) -- (-3.25,0);
  \node[tick, anchor=west, text=ink] at (-3.13,0) {reference (pretrained / trained)};
  \draw[edit, curve] (.9,0) -- (1.35,0);
  \node[tick, anchor=west, text=ink] at (1.47,0) {written by \textcolor{astra}{\textbf{Astra}}};
\end{scope}
\end{tikzpicture}
    }
    \vspace{-0.5em}
    \caption{
    \textbf{(A) Construct:} Astra writes nanoTabPFN weights from scratch, achieving competitive performance relative to a pretrained baseline;
    \textbf{(B) Repair:} Astra programs a corrective computation into pretrained TabPFN-3.5 weights, resolving a signed-product regression failure;
    \textbf{(C) Steer:} Astra translates a natural-language instruction into a reusable KV-cache editing program that changes a fitted model's predictions while keeping its weights frozen.
    }
    \label{fig:hero-plot}
\end{figure}

\newpage
\section{Introduction}
\label{sec:intro}

Neural networks are typically programmed indirectly: rather than specifying the computations they should perform, we optimize their parameters on data and rely on the resulting weights to implement useful algorithms. This paradigm was recently challenged by \emph{weight-space programming (WSP)}, which shows that large language models (LLMs) can reason about neural-network weights and directly construct or modify them~\citep{takehi2026weightspace}.

Tabular foundation models (TFMs) offer a particularly interesting setting for WSP. TFMs such as TabPFN predict on unseen datasets by conditioning on labeled examples, without updating the TFM's weights~\citep{hollmann2023tabpfntransformersolvessmall,hollmann2025tabpfn,tabpfn35}. Unlike task-specific predictors, their parameters must therefore implement reusable statistical computations that adapt to different datasets. However, pretraining is costly: it requires generating data from synthetic priors and running multi-stage training across a large space of design choices, making it expensive in both time and compute.

These costs motivate investigating whether statistical computation can be programmed directly into TFM architectures, rather than learned.
But programming a TFM need not start from scratch. A pretrained model may already contain useful statistical information, yet fail to exploit it for a particular task.
Instead of fine-tuning~\citep{thomas2024localpfn, rubachev2025finetuning, buehler2026causaldataaugmentationrobust} or selecting context~\citep{feuer2024tunetables, thomas2024localpfn}, directly editing its weights could repair such failures while preserving its broader predictive capabilities.
Furthermore, we can edit the KV-cache of TFMs that store forward-pass activations, representing the TFM's fitted prediction model.
Editing the KV-cache can steer predictions while leaving the weights untouched.

Motivated by these possibilities, we investigate \textbf{three forms of LLM-driven programming of TFMs: constructing, repairing, and steering.} We use GPT-6 Astra~\citep{openai2026astra} to investigate these three settings, while explicitly prohibiting gradient-based optimization throughout. Our contributions are:
\begin{enumerate}[label=(\Alph*), leftmargin=\dimexpr\leftmargini-1.5em\relax]
    \item \textbf{Constructing a tabular foundation model from zero weights.}
    We unlock weight-space programming for TFMs. Using Astra to program nanoTabPFN~\citep{pfefferle2025nanotabpfn}, we achieve a mean ROC AUC of $0.849$ across $20$ TabArena datasets~\citep{erickson2025tabarena}, compared with $0.876$ for a pretrained baseline.

    \item \textbf{Repairing a predictive failure in TabPFN-3.5 in the weight space.}
    We identify a failure of TabPFN-3.5 on signed-product regression tasks and fix it through weight-space programming. Astra can diagnose and repair it by directly modifying TabPFN-3.5's weights. The intervention restores near-perfect predictions on held-out synthetic datasets, with only a $0.08\%$ average relative increase in error on TabArena-Lite.

    \item \textbf{Steering TabPFN-3.5 through KV-cache programming.}
    We demonstrate that Astra can translate natural-language instructions into reusable KV-cache editing programs. Without modifying TabPFN-3.5's weights, these programs reverse linear trends and approximate sine-to-square-wave transformations on unseen synthetic datasets.
\end{enumerate}

Our findings provide initial evidence that LLM-driven programming could complement gradient-based optimization across TFMs' development, from pretraining to post-training.

\section{Related Work}
\label{sec:related}
\textbf{Tabular foundation models (TFMs).}\quad
TFMs learn reusable prediction procedures through pretraining on synthetic datasets and adapt to new datasets through in-context inference~\citep{hollmann2023tabpfntransformersolvessmall, qu2025tabicltabularfoundationmodel, hollmann2025tabpfn, qu2026tabiclv2, tabpfn35}. Methods such as TuneTables~\citep{feuer2024tunetables} further explore optimizing context representations for pretrained models. In contrast, we investigate whether LLMs can explicitly construct or modify the computations performed by TFMs, through either their shared weights or dataset-specific KV-caches.
For this work, we focus our investigation on TabPFN-3.5~\citep{tabpfn35} and nanoTabPFN~\citep{pfefferle2025nanotabpfn}, a lightweight, educational implementation of TabPFN.

\textbf{Editing and steering.}\quad
Prior work changes trained models without retraining, either by editing weights, as in model
editing~\citep{meng2022rome,meng2023memit} and task arithmetic~\citep{ilharco2023taskarithmetic}, or by editing internal states, as in activation steering~\citep{turner2023actadd,zou2023repe}, cache steering~\citep{belitsky2025cachesteering}, and prefix tuning~\citep{li2021prefix}. These edits are computed from optimization or activation statistics. Closest to our work, weight-space programming (WSP) shows that an LLM can read and write neural network weights directly, without optimization, on control policies and LLMs~\citep{takehi2026weightspace}. We extend this idea to TFMs: the instruction is stated in natural language, and an LLM programs either the model's weights or its KV cache by reasoning about the computation it performs.

\textbf{Interpreting neural networks.}\quad
Prior work investigates the internal computations of TabPFN and related PFNs through activation patching and representation analysis, identifying functional roles of attention heads and information encoded in hidden states~\citep{gupta2026wherecomputation,sharma2026spectral,gupta2026lookingglass}.
Automated interpretability agents further demonstrate how LLMs can investigate neural-network components by formulating and testing hypotheses~\citep{shaham2024maia}. Building on these directions, we design our setups such that the LLM agent must reason about the TFM's internals before programming them, and we instruct it to keep interpretable traces. These traces make the results auditable and every solution explainable.

\section{Methodology and Experiments}
\label{sec:method}
We investigate three forms of LLM-driven programming of TFMs in separate experimental settings (see Figure~\ref{fig:hero-plot}).
We instruct Astra to program different components of the TFMs. We detail each experimental setting in the respective subsections. Across all settings, we prohibit Astra from doing gradient-based optimization and restrict access to the hold-out evaluation datasets. We audited the recorded transcripts and programs for compliance with these rules and found no violations.

\subsection{Weight-space Programming nanoTabPFN}

Tabular foundation models typically acquire their predictive capabilities through pretraining on synthetic datasets. In this experiment, we study whether an LLM can instead construct a useful predictor by directly specifying the weights of a zero-initialized TFM.
We then task Astra to weight-space program nanoTabPFN~\citep{pfefferle2025nanotabpfn}, where we deliberately provide no data and forbid any forward passes. Then, we evaluate the resulting model on TabArena datasets.

\label{sec:exp-wsp}
\textbf{Experimental setup.}\quad
Following \citet{takehi2026weightspace}, we provide Astra with nanoTabPFN's 3.7 million zero-initialized weights, source code and preprocessing, a specification of the forward pass, and the evaluation protocol.
The agent has no access to datasets or pretrained weights.
It may write weights through code but cannot fit parameters to data, compute gradients, run an optimizer, or execute or simulate the network.
We execute each constructed network only at final evaluation.
We evaluate mean ROC AUC and log loss on 20 TabArena-v0.1~\citep{erickson2025tabarena} classification datasets with at most 10,000 rows, 100 features, and 10 classes.
As a baseline, we compare to a nanoTabPFN pretrained on 160,000 synthetic tables from TFM-Playground~\citep {pfefferle2026tfmplayground} prior.
Appendix~\ref{app:setup} provides the full protocol.

\textbf{Results.}\quad
The resulting model achieves a mean ROC AUC of $0.849$ and log loss of $0.397$ on the 20 datasets, compared to $0.876$ and $0.312$ for the pretrained baseline.
Notably, only $0.42\%$ of the weights were filled in, which can be indicative of overparametrization of nanoTabPFN.
Achieving such comparable performance to the pretrained baseline demonstrates that an LLM can directly express an interpretable algorithm through the weights of a TFM architecture.

\textbf{Programmed Algorithm.}\quad
Rather than attempting to reproduce pretrained weights, Astra implements a multiscale kernel naive Bayes classifier~\citep{john1995estimating} by re-wiring nanoTabPFN's attention and MLP operations.
Datapoint attention compares examples within each feature column and uses Gaussian kernels to estimate per-feature class probabilities.
Feature attention then combines their log evidence, weighted by feature predictiveness, while a later block calibrates predictions using neighboring training rows. To control \texttt{LayerNorm}, Astra introduces large constant coordinates that make normalization act approximately as a fixed rescaling.
Such construction assigns an explicit computational purpose to the written weights, making the intended algorithm interpretable (Appendix~\ref{app:setup}).

\subsection{Repairing Signed-Product Prediction Failure in TabPFN-3.5}
\label{sec:exp-failure}

Our analysis found that TabPFN-3.5~\citep{tabpfn35} fails on signed-product regression once the product has more than four input features by predicting values near zero for almost every query.
We investigate whether Astra can diagnose this failure and then program a corrective computation into the pretrained TabPFN-3.5 weights, while preserving its predictive performance on TabArena-Lite.

\textbf{Experimental setup.}\quad
We study noiseless signed-product regression,
\vspace{-4pt}
\begin{equation*}
  y = \prod_{i=1}^{k} s_i \exp(u_i), \quad s_i \in \{-1,+1\}, \quad u_i \sim \mathcal{U}(-0.5,0.5)
\end{equation*}

\vspace{-10pt}

with $k \in \{5,6\}$ inputs and $N \in \{256,512\}$ context rows using pretrained TabPFN-3.5.
Astra receives the model source, weights, and labeled training data.
Additionally, Astra receives 15 OpenML datasets and may request aggregate feedback on hidden validation datasets and TabArena-Lite inner splits. It may inspect activations and edit weights using closed-form fits, but cannot perform gradient-based optimization. We evaluate the repaired model on 240 held-out signed-product datasets and the TabArena-Lite outer test splits. Appendix~\ref{app:signed-products} provides the full protocol.

\begin{figure}[t]
  \centering
  \begin{minipage}[c]{0.59\linewidth}
    \centering
    \providecommand{\spdensity}[6]{%
  \begin{scope}[shift={(#1,#2)}]
    \fill[#3!18] plot[domain=-1.05:1.05, samples=70]
      (\x,{#4*exp(-((\x+0.6)/0.17)^2) + #5*exp(-((\x-0.6)/0.17)^2)}) -- (1.05,0) -- (-1.05,0) -- cycle;
    \draw[#3, line width=0.6pt] plot[domain=-1.05:1.05, samples=70]
      (\x,{#4*exp(-((\x+0.6)/0.17)^2) + #5*exp(-((\x-0.6)/0.17)^2)});
    \draw[wspinktwo, line width=0.4pt] (-1.1,0) -- (1.1,0);
    \foreach \t in {-0.6,0,0.6} \draw[wspinktwo, line width=0.4pt] (\t,0) -- (\t,-0.05);
    \draw[wspink, densely dashed, line width=0.5pt] (#6,0) -- (#6,0.6);
  \end{scope}}

\begin{tikzpicture}[
    font=\figurefont, text=wspink,
    flow/.style={-{Stealth[length=3.5pt,width=3.2pt]}, line width=0.55pt, draw=#1},
    flow/.default=wspinktwo,
    box/.style={line width=0.55pt, rounded corners=2pt, minimum height=0.72cm, minimum width=1.35cm,
      inner sep=2pt, align=center},
    model/.style={box, draw=wspinktwo, fill=wspgrid},
    lab/.style={anchor=north, inner sep=1pt, align=center, font=\figurefont},
  ]
  \node[model] (early) at (0.7,0) {Early\\layers};
  \node[model] (later) at (2.55,0) {Later\\layers};
  \node[model] (head)  at (4.4,0) {Regression\\head};
  \draw[flow] (early.east) -- (later.west);
  \draw[flow] (later.east) -- (head.west);

  \node[lab] at (0.7,-0.5) {count of negative\\inputs is decodable};
  \node[lab, text=wspafter] at (2.55,-0.5) {gate closes on\\other tasks};
  \node[lab, text=wspafter] at (4.4,-0.5) {parity selects\\the sign};

  \node[box, draw=wspafter, fill=wspafter!12, minimum height=0.36cm, minimum width=0.7cm] (gate) at (3.2,0.95) {gate};
  \draw[wspafter, line width=0.55pt] (early.north) |- (gate.west);
  \draw[flow=wspafter] (gate.east) -| (head.north);
  \node[anchor=south, inner sep=1.5pt, text=wspafter] at (1.7,0.95) {carry the count};

  \draw[flow=wspbefore] (head.east) -- ++(0.25,0) |- (5.6,0.6);
  \draw[flow=wspafter]  (head.east) -- ++(0.25,0) |- (5.6,-0.95);
  \spdensity{6.8}{0.4}{wspbefore}{0.5}{0.5}{0}
  \spdensity{6.8}{-1.15}{wspafter}{0}{0.5}{0.6}
  \node[anchor=south west, inner sep=1pt, text=wspbefore, font=\figurefont\bfseries] at (5.7,1.08) {Pretrained};
  \node[anchor=south west, inner sep=1pt, text=wspafter, font=\figurefont\bfseries] at (5.7,-0.47) {Edited};
  \foreach \t/\l in {-0.6/$-y$, 0/$0$, 0.6/$y$}
    \node[anchor=north, inner sep=1.5pt] at ({6.8+\t},-1.2) {\l};
  \path[use as bounding box] (-0.3,-2.37) rectangle (7.95,2.41);
  \node[anchor=north west, inner sep=0pt, font=\figurefont\bfseries] at (-0.3,2.27) {(a)};
\end{tikzpicture}
  \end{minipage}\hfill
  \begin{minipage}[c]{0.31\linewidth}
    \centering
    \includegraphics[width=\linewidth]{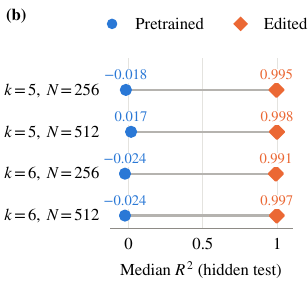}
  \end{minipage}
  \vspace{-2em}
  \caption{Signed-product intervention. \textbf{(a)}~Schematic of the diagnosis and the edit, for a query with
    target $y>0$. The pretrained model places predictive mass at both $-y$ and $y$, so its mean (dashed) is near zero, while
    its early layers already encode the number of negative inputs. The edit (orange) carries this count to the
    regression head, where its parity selects the sign. \textbf{(b)}~Median $R^2$ over 60 hidden datasets per regime
    ($k$ inputs, $N$ context rows, 500 query rows), before and after the edit.}
  \label{fig:signed-product-results}
\end{figure}
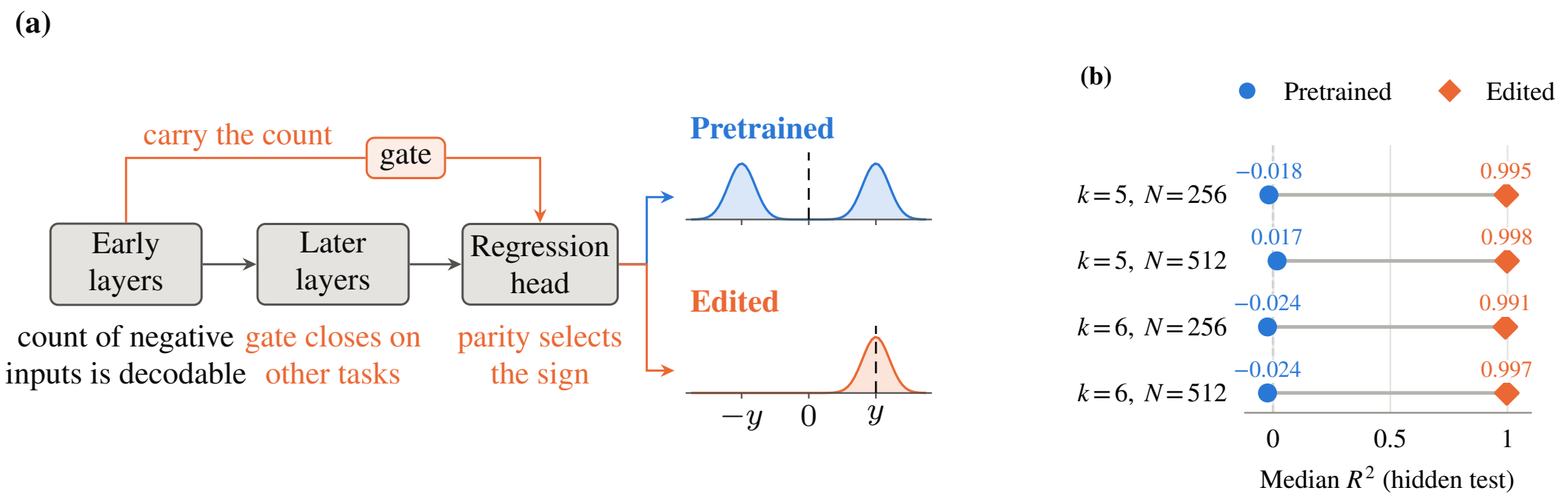

\textbf{Astra's Diagnosis and Repair.}\quad
In our experiment, Astra diagnoses the following:
\begin{enumerate*}[label=(\roman*)]
    \item TabPFN-3.5 fails on mixed-signed products: TabPFN-3.5 accurately predicts products when all feature values are positive but fails when negative values are introduced;
    \item TabPFN-3.5 correctly predicts the magnitude: the predictive distribution is bimodal with modes near $+|y|$ and $-|y|$ that cancel in the mean (\cref{fig:product-bimodal}). Moving this mass to the true sign restores accurate predictions, so choosing the sign is the main obstacle.
    \item The sign information is present early but lost later: The sign of the product is the parity of the number of negative inputs, and linear probes recover this count from early representations but not from late ones.
\end{enumerate*}
With this evidence, Astra repairs TabPFN-3.5 by constructing a circuit that carries the count to the prediction head and selects the sign from its parity, with an internal gate that limits interference on other tasks.

\textbf{Results.}\quad
Across the four experimental settings, the repair raises median per-dataset $R^2$ from $-0.024$--$0.017$ to $0.991$--$0.998$ on held-out synthetic datasets (Figure~\ref{fig:signed-product-results}b).
Despite modifying the weights, the performance on TabArena-Lite also remains nearly unchanged: the geometric-mean edited-to-pretrained error ratio is 1.0008, corresponding to an aggregate relative error increase of just $0.08\%$ (Appendix~\ref{app:signed-products}, Table~\ref{tab:product-preservation}).
These results demonstrate that Astra can repair a specific predictive failure through weight-space programming while preserving the model's broader predictive capabilities. Appendix~\ref{app:signed-products} provides the full diagnostic evidence, repair details, and evaluation results.

\subsection{Steering TabPFN-3.5 through KV-Cache Programming}
\label{sec:exp-kv}

Our first two experiments directly program model weights. However, a TFM's predictions also depend on the labeled context of each dataset. In TabPFN-3.5, the keys and values associated with context rows can be stored in a KV-cache and reused to predict new rows.
This motivates further exploration into steering TFMs' predictions: given a description in natural language, can an LLM program the KV-cache to steer TFMs' predictions without modifying its weights?

We investigate this by asking Astra to write reusable cache-editing programs for instructions such as "flip the slope". Rather than editing a particular dataset's cache manually, Astra must produce a program that operates on the caches of previously unseen datasets.

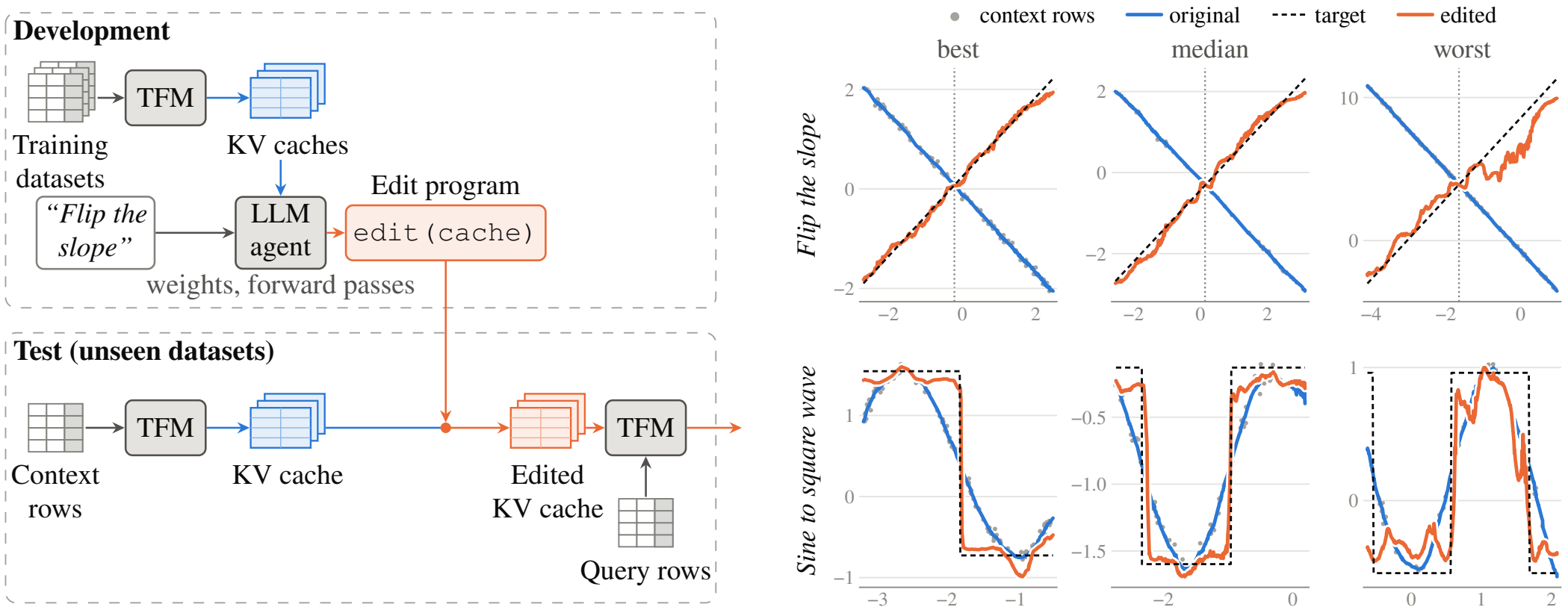
\begin{figure}[t]
    \centering
    \begin{minipage}{0.49\textwidth}
        \centering
        \resizebox{\linewidth}{!}{%
            \providecommand{\kvtable}[2]{%
  \begin{scope}[shift={({#1-0.27},{#2-0.24})}]
    \foreach \col in {0,1,2}
      \foreach \row in {0,...,3}
        \draw[datastroke, fill=white, line width=0.4pt] ({0.18*\col},{0.12*\row}) rectangle ++(0.18,0.12);
    \foreach \row in {0,...,3}
      \draw[datastroke, fill=wspdata!40, line width=0.4pt] (0.36,{0.12*\row}) rectangle ++(0.18,0.12);
  \end{scope}}

\providecommand{\kvsheets}[3]{%
  \begin{scope}[shift={(#1,#2)}]
    \foreach \layer in {2,1}
      \draw[#3, fill=#3!12, line width=0.5pt]
        ({-0.3+0.07*\layer},{-0.2+0.07*\layer}) rectangle ++(0.6,0.4);
    \draw[#3, fill=#3!12, line width=0.5pt] (-0.3,-0.2) rectangle (0.3,0.2);
    \foreach \row in {1,2,3}
      \draw[#3!55, line width=0.3pt] (-0.3,{-0.2+0.1*\row}) -- (0.3,{-0.2+0.1*\row});
    \draw[#3!55, line width=0.3pt] (0,-0.2) -- (0,0.2);
  \end{scope}}

\begin{tikzpicture}[
    font=\figurefont, text=wspink,
    datastroke/.style={draw=wspdata!60!wspinktwo},
    flow/.style={-{Stealth[length=3.5pt,width=3.2pt]}, line width=0.55pt, draw=#1},
    flow/.default=wspinktwo,
    box/.style={line width=0.55pt, rounded corners=2pt, minimum height=0.55cm, inner sep=2pt, align=center},
    model/.style={box, draw=wspinktwo, fill=wspgrid, minimum width=0.8cm},
    lab/.style={anchor=north, inner sep=1pt, align=center, font=\figurefont},
    phase/.style={draw=wspdata, dashed, line width=0.4pt, rounded corners=3pt},
    phaselab/.style={anchor=north west, inner sep=2pt, font=\figurefont\bfseries},
  ]
  \def\xa{0.35} \def\xb{1.45} \def\xc{2.6} \def\xd{4.25} \def\xe{5.2} \def\xf{6.25}

  \draw[phase] (-0.15,0.85) rectangle (6.95,-2.1);
  \node[phaselab] at (-0.15,0.85) {Development};

  \kvtable{\xa+0.12}{0.12} \kvtable{\xa+0.06}{0.06} \kvtable{\xa}{0}
  \node[lab] at (\xa+0.05,-0.33) {Training\\datasets};
  \node[model] (tfmA) at (\xb,0) {TFM};
  \draw[flow] (\xa+0.42,0) -- (tfmA.west);
  \kvsheets{\xc}{0}{wspbefore}
  \node[lab] at (\xc+0.07,-0.33) {KV caches};
  \draw[flow=wspbefore] (tfmA.east) -- (\xc-0.31,0);

  \node[model, minimum width=0.9cm] (agent) at (\xc,-1.35) {LLM\\agent};
  \node[box, datastroke, fill=white, font=\figurefont\itshape] (instr) at (0.75,-1.35) {``Flip the\\slope''};
  \draw[flow] (instr.east) -- (agent.west);
  \draw[flow=wspbefore] (\xc,-0.62) -- (agent.north);
  \node[lab, text=wspinktwo] at (\xc,-1.72) {weights, forward passes};
  \node[box, draw=wspafter, fill=wspafter!12, font=\figurefont\ttfamily] (prog) at (\xd,-1.35) {edit(cache)};
  \draw[flow=wspafter] (agent.east) -- (prog.west);
  \node[lab, anchor=south] at (prog.north) {Edit program};

  \draw[phase] (-0.15,-2.35) rectangle (6.95,-5.05);
  \node[phaselab] at (-0.15,-2.35) {Test (unseen datasets)};

  \def\yt{-3.3}
  \kvtable{\xa}{\yt}
  \node[lab] at (\xa,\yt-0.33) {Context\\rows};
  \node[model] (tfmB) at (\xb,\yt) {TFM};
  \draw[flow] (\xa+0.3,\yt) -- (tfmB.west);
  \kvsheets{\xc}{\yt}{wspbefore}
  \node[lab] at (\xc+0.07,\yt-0.33) {KV cache};
  \draw[flow=wspbefore] (tfmB.east) -- (\xc-0.31,\yt);

  \coordinate (dot) at (\xd,\yt);
  \draw[line width=0.55pt, wspbefore] (\xc+0.45,\yt) -- (dot);
  \draw[flow=wspafter] (dot) -- (\xe-0.31,\yt);
  \draw[flow=wspafter, shorten >=1.5pt] (prog.south -| dot) -- (dot);
  \fill[wspafter] (dot) circle (1.5pt);

  \kvsheets{\xe}{\yt}{wspafter}
  \node[lab] at (\xe+0.07,\yt-0.33) {Edited\\KV cache};
  \node[model] (tfmC) at (\xf,\yt) {TFM};
  \draw[flow=wspafter] (\xe+0.45,\yt) -- (tfmC.west);
  \draw[flow=wspafter] (tfmC.east) -- ++(0.55,0);
  \kvtable{\xf}{\yt-0.95}
  \node[lab] at (\xf,\yt-1.28) {Query rows};
  \draw[flow] (\xf,\yt-0.69) -- (tfmC.south);
\end{tikzpicture}
        }
    \end{minipage}
    \hfill
    \begin{minipage}{0.49\textwidth}
        \centering
        \includegraphics[width=\linewidth]{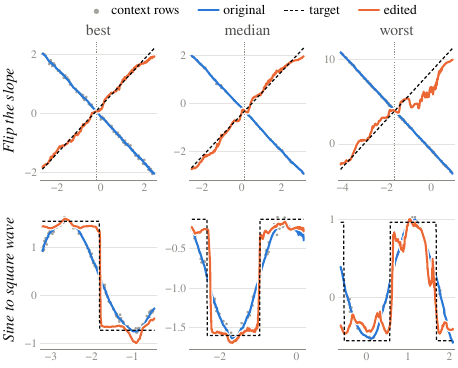}
    \end{minipage}
    \caption{Programming the KV-cache of TabPFN-3.5. Left: during development, Astra reads the KV-caches of training datasets and writes an edit program;
    in the test phase, the program edits the KV-cache of an unseen dataset, and the TFM predicts the query rows from the edited cache.
    Right: the best, median, and worst of test datasets for \emph{flip-the-slope} (top) and \emph{sine-to-square-wave} (bottom).}
    \label{fig:kv-cache}
\end{figure}

\paragraph{Experimental Setup.}
Figure~\ref{fig:kv-cache} (left) shows the development and test phases.
We consider two families of synthetic one-feature regression tasks: \emph{flip-the-slope} and \emph{sine-to-square-wave}.
During development, Astra receives a natural-language instruction and example datasets.
It then independently explores candidate interventions by inspecting the model's weights and internal representations, writing cache-editing programs, and evaluating their effects through forward passes.
This iterative exploration proceeds without human intervention or gradient-based optimization.
Once satisfied, Astra submits an \texttt{edit(cache)} function in Python, which is evaluated on ten unseen datasets from the same family.
Storing weights or precomputed artifacts is prohibited to ensure that the function derives its interventions from each dataset's cache rather than memorized solutions.
Appendix~\ref{app:kvp} provides the complete setup.

\textbf{Results.}\quad
Figure~\ref{fig:kv-cache} (right) shows the steered predictions on unseen test datasets; \cref{fig:kvp-all} in \cref{app:kvp} shows all twenty.
Given the natural-language instructions, Astra took about $20$ minutes to write the slope program and $70$ minutes to write the square-wave program.
Both programs approximate the intended transformations on previously unseen datasets, although the quality of the interventions varies across datasets.
These results demonstrate that reusable programs can steer predictions through KV-cache interventions.

\textbf{How the Programs Work.}\quad
The two programs follow different approaches.
For \emph{flip-the-slope}, Astra estimates the positions of context rows from variation in early-layer values, fits the later-layer values as functions of those positions, and reflects the fitted functions. For \emph{sine-to-square-wave}, the program orders context rows using key similarities, estimates their positions along the sinusoid, and moves cached values toward those associated with peaks or troughs.
Further details in Appendix~\ref{app:kvp-edits}.

\section{Discussion}
\label{sec:discussion}

This work provides initial evidence that LLMs can directly program useful computations in tabular foundation models, beyond conventional gradient-based optimization.
Our experiments illustrate distinct forms of programming.
In the first, Astra translates interpretable algorithms into the operations of a transformer.
In the second, it identifies sign-relevant information already present in pretrained representations and constructs a parity-based correction to make that information useful for prediction.
In the third, it writes programs that steer the TabPFN-3.5's predictions without changing its weights.
Although the three settings pursue different objectives, together they demonstrate that LLMs can intervene in both reusable model computations and fitted representations.

These findings remain exploratory, but point toward a broader paradigm in which LLM-driven programming complements gradient-based optimization across TFM development and usage.
Beyond improving predictive behavior, this framework could also provide a more interpretable approach to understanding and developing TFMs.

\newpage
\section*{Acknowledgements}
Funded by the European Union. Views and opinions expressed are, however, those of the author(s) only and do not necessarily reflect those of the European Union or the European Commission. Neither the European Union nor the European Commission can be held responsible for them. This work was supported by the European Union's Horizon Europe research and innovation program under grant agreement No 101214398 (ELLIOT).
L.P. acknowledges funding by the Deutsche Forschungsgemeinschaft (DFG, German Research Foundation) under SFB 1597 (SmallData), grant number 499552394;
F.H. acknowledges the financial support of the Hector Foundation.
  \begin{center}
    \begin{minipage}{0.2\textwidth}
      \centering
      \includegraphics[width=\textwidth]{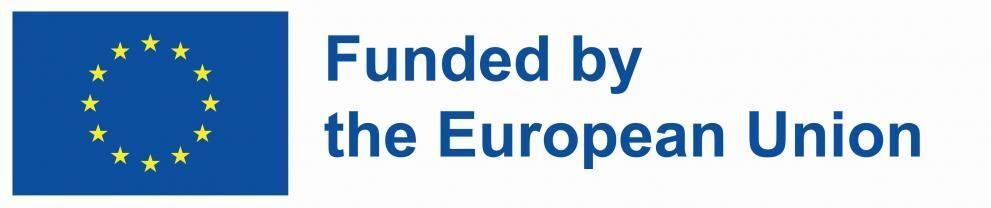}
    \end{minipage}
    \begin{minipage}{0.2\textwidth}
      \centering
      \includegraphics[width=\textwidth]{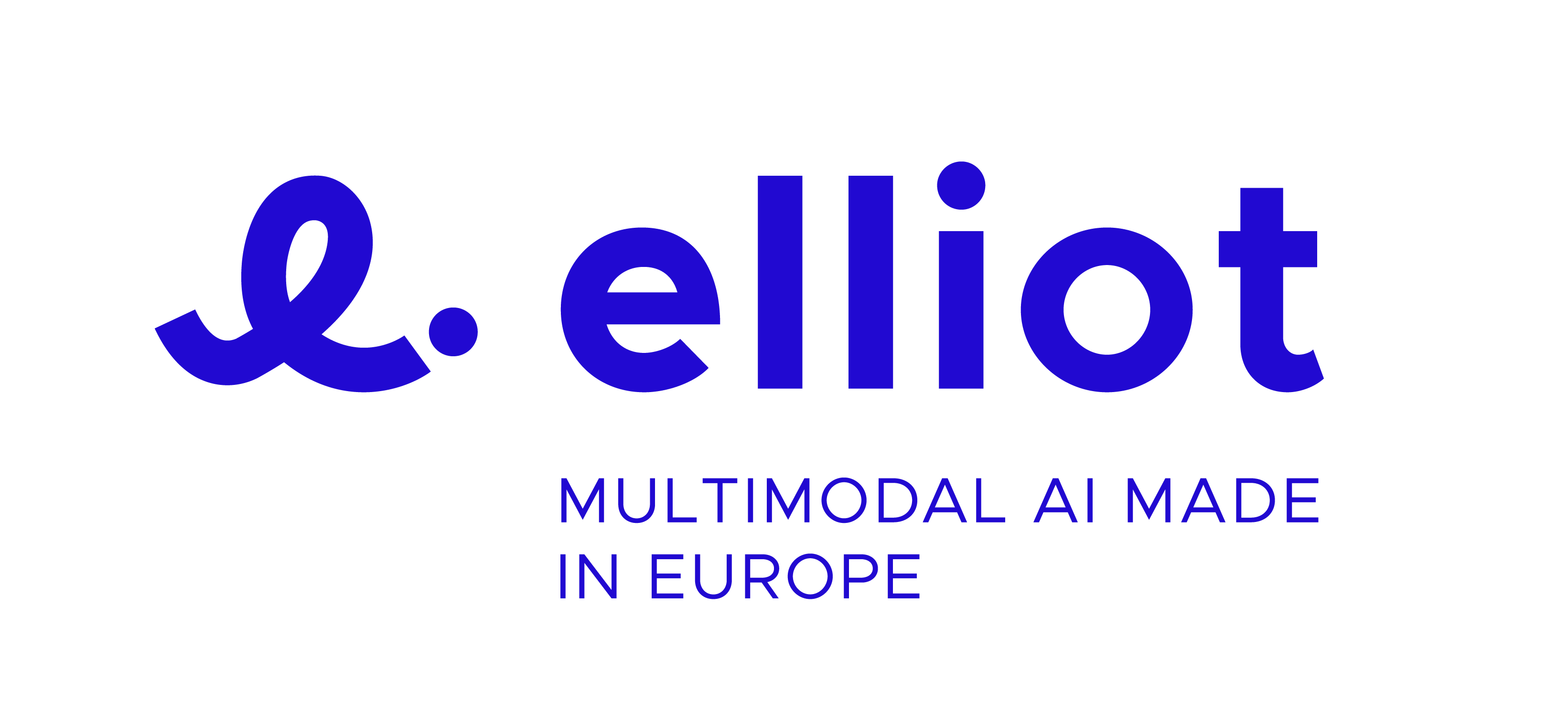}
    \end{minipage}
  \end{center}

\bibliography{references}
\bibliographystyle{unsrtnat}

\clearpage
\appendix
\section{Writing nanoTabPFN from zero weights: full setup}
\label{app:setup}

This appendix documents the experimental protocol, agent prompt, construction, and evaluation of WSP-nanoTabPFN.

\paragraph{Network.}
nanoTabPFN is the \texttt{NanoTabPFNModel} of TFM-Playground at commit \texttt{13a1df3}, in its default classifier
configuration: embedding width 192, 6 attention heads of width 32, MLP width 768, 6 blocks and 10 outputs, which gives
116 tensors and 3,717,514 weights. Every weight is zero at the start. Each block applies attention between the cells
of a row, then attention between the rows of a column (keys and values from the training rows only), then an MLP,
each followed by a residual connection and a LayerNorm. The decoder maps the target cell of each test row to 10
logits, of which the first $C$ are used. Before the network, TFM-Playground's \texttt{TabularClassifier} preprocesses
each table on its training rows: it drops constant columns, imputes missing values and adds an indicator column for
each column that has them, ordinal-encodes categorical columns and standardises every column. The architecture and
the preprocessing are fixed; only the weights are set.

\paragraph{Agent.}
The agent is GPT-6 Astra (\texttt{gpt-6-astra}) through the Codex CLI. Each turn is one call of \texttt{codex exec}
(first turn) or \texttt{codex exec resume} (later turns), with the harness's message on standard input, reasoning
effort \texttt{xhigh}, reasoning summaries \texttt{detailed} and verbosity \texttt{medium}. Every Codex tool is switched
off, among them the shell, code execution, web search, MCP servers, plugins, sub-agents and the image and browser
tools, and Codex's system instructions are replaced by the single line ``Answer the user's request.'' Astra therefore
acts only through the text of its replies. The harness checks the event stream of every turn and stops the session if
it holds anything other than messages and reasoning. Every turn passed this check.

\paragraph{Interaction.}
The first message is the prompt in \cref{lst:nano-prompt}. Each reply may contain one block that starts with
\verb|```python|. The harness runs that block in the sandbox and sends its printed output back as the next message.
A block preceded by fewer than 25 words of explanation is returned unrun and not counted, at most twice in a row, and
a reply with more than one block runs nothing. A reply without a block ends the session; it is Astra's final
description of its network. The limits are 120 Python runs, 900\,s and 64\,GB per run, 20,000 characters of
returned output per run, 400 turns and 4\,h per model call. The harness sends these messages, verbatim:
\begin{itemize}
  \item after a run: \texttt{OUTPUT of Python run <k> of at most 120:}, followed by the output in a code block;
  \item after a block without explanation: \texttt{EXPLAIN EVERY STEP: your reply had no explanation before its block,
    so the block was not run (it did not count). Reply again with the plain-English explanation the task asks for,
    followed by the block.};
  \item after several blocks: \texttt{Your reply contained more than one python / practice / probe block; nothing was
    run. Reply with exactly one block.};
  \item after the last run: \texttt{All 120 Python runs are used; the code was not run.}
\end{itemize}

\paragraph{Sandbox.}
Each run is a fresh Python 3.12 process in new Linux user, mount, network and PID namespaces (\texttt{unshare
-rmnpf}), with its root changed to a directory that holds only the system libraries, the Python installation, a
read-only virtual environment and the session's working folder \texttt{/work}. The sandbox has no network and no GPU
and uses 8 CPU threads. Only files in \texttt{/work} persist between runs. \texttt{/work/source} holds the source of
the network and of the preprocessing (\texttt{nanotabpfn.py}, \texttt{interface.py}); no dataset, prior or trained
weight is reachable. Before each run the harness defines \texttt{SHAPES}, \texttt{NAMES}, \texttt{load(name)},
\texttt{save(name, array)} and \texttt{saved()}. \texttt{save} accepts only a finite array of the tensor's shape and
stores it in float32. The saved tensors are the network; tensors never saved stay zero. The available packages are
NumPy, SciPy, scikit-learn, PyTorch (CPU), pandas, SymPy and NetworkX.

\paragraph{Rules and audit.}
The prompt forbids (a) training of any kind, including fitting weights to data or labels, whether practice data or
tables the agent generates, (b) implementing or simulating the network's computation, or any part of it, on any input,
and (c) internet access. It allows writing the weights by code from the agent's own design. The sandbox enforces
(c); we check (a) and (b) by reading every transcript and program (\cref{sec:method}). The final description follows
a template: the classifier the network implements, the layout of the 192 dimensions of a cell and why the norms
preserve it, what each attention, each MLP and the decoder compute, the main numerical choices (scales,
temperatures, kernel widths, priors), and where the network should be weakest.

\paragraph{Evaluation.}
After a session ends, the harness scores its network once on the 20 datasets of \cref{tab:nano-datasets}: the
classification tasks of TabArena-v0.1 with at most 10,000 rows, 100 features and 10 classes. On each dataset the
context is the training part of TabArena's first split (fold 0, repeat 0; two thirds of the rows, at most 6,666), and
the network predicts every test row. Predictions come from TFM-Playground's \texttt{get\_openml\_predictions} with
\texttt{TabularClassifier}, on CPU in float32. Per dataset we compute the one-vs-rest ROC AUC (for binary tasks, from
the probability of the positive class) and the log loss with probabilities clipped at $10^{-15}$, and we report
their unweighted means over the 20 datasets (per-dataset values in \cref{tab:nano-per-dataset}). Scoring one network takes about 3 minutes. The prompt gives
the datasets' feature counts as 4 to 85; by OpenML's metadata, two datasets have more (86 and 94).

\begin{table}[ht]
  \centering
  \caption{The 20 evaluation datasets (TabArena-v0.1). Rows cover both parts of the split; features are the raw
    columns before preprocessing, from OpenML's metadata.}
  \label{tab:nano-datasets}
  \small
  \begin{tabular}{lrrrr}
    \toprule
    Dataset & OpenML task & Rows & Features & Classes \\
    \midrule
    anneal                                 & 363614 &    898 & 38 & 5 \\
    Bank\_Customer\_Churn                  & 363619 & 10,000 & 10 & 2 \\
    blood-transfusion-service-center       & 363621 &    748 &  4 & 2 \\
    churn                                  & 363623 &  5,000 & 19 & 2 \\
    coil2000\_insurance\_policies          & 363624 &  9,822 & 85 & 2 \\
    credit-g                               & 363626 &  1,000 & 20 & 2 \\
    diabetes                               & 363629 &    768 &  8 & 2 \\
    Fitness\_Club                          & 363671 &  1,500 &  6 & 2 \\
    hazelnut-spread-contaminant-detection  & 363674 &  2,400 & 30 & 2 \\
    Is-this-a-good-customer                & 363682 &  1,723 & 13 & 2 \\
    Marketing\_Campaign                    & 363684 &  2,240 & 25 & 2 \\
    maternal\_health\_risk                 & 363685 &  1,014 &  6 & 3 \\
    NATICUSdroid                           & 363689 &  7,491 & 86 & 2 \\
    polish\_companies\_bankruptcy          & 363694 &  5,910 & 64 & 2 \\
    qsar-biodeg                            & 363696 &  1,054 & 41 & 2 \\
    seismic-bumps                          & 363700 &  2,584 & 15 & 2 \\
    splice                                 & 363702 &  3,190 & 60 & 3 \\
    students\_dropout\_and\_academic\_success & 363704 &  4,424 & 36 & 3 \\
    taiwanese\_bankruptcy\_prediction      & 363706 &  6,819 & 94 & 2 \\
    website\_phishing                      & 363707 &  1,353 &  9 & 3 \\
    \bottomrule
  \end{tabular}
\end{table}

\paragraph{Pretrained nanoTabPFN baseline.}
We use nanoTabPFN in the same architecture and
configuration as the programmed model, trained
from random initialization on TFM-Playground's
default nanoTabICL prior with its ExtraTrees filter.
Training uses one seed and 20,000 steps with eight
synthetic tables per step (160,000 tables in total),
a learning rate of $10^{-3}$, and 1,000 warm-up steps.
We evaluate the final checkpoint with the same
inference code and protocol as the programmed model.

\paragraph{Hardware and software.}
The session ran as one SLURM job on a CPU node (AMD EPYC 9655; 32 cores and 128\,GB reserved; no GPU), which ran
both the harness and the sandbox; Astra itself ran on OpenAI's servers. The session ran on 4~November~2026.
Software: Python 3.12, PyTorch 2.9.0, NumPy 2.5.4, scikit-learn 1.9.1, openml 0.15.1, TFM-Playground at commit
\texttt{13a1df3}, Codex CLI 0.164.0.

\paragraph{Session.}
We ran two sessions under this protocol and report the better one (\cref{tab:nano-sessions}). Astra's first block
came without an explanation; the harness returned it, and Astra resent it with one. Every other block ran.

\begin{table}[ht]
  \centering
  \caption{The session that wrote WSP-NanoTabPFN. Scores are means over the 20 datasets of
    \cref{tab:nano-datasets}.}
  \label{tab:nano-sessions}
  \small
  \begin{tabular}{lr}
    \toprule
    & WSP-NanoTabPFN \\
    \midrule
    Turns                                    & 7      \\
    Python runs                              & 5      \\
    Blocks returned for lack of explanation  & 1      \\
    Wall-clock time (min)                    & 20     \\
    Nonzero weights                          & 15,513 (0.42\%) \\
    ROC AUC                                  & 0.849  \\
    Log loss                                 & 0.397  \\
    \bottomrule
  \end{tabular}
\end{table}

\begin{table}[ht]
  \centering
  \caption{Per-dataset ROC AUC and log loss on TabArena's first split of WSP-NanoTabPFN (WSP) and the SGD Baseline
    (SGD). WSP-NanoTabPFN has the higher AUC on 3 of the 20 datasets (diabetes, Is-this-a-good-customer, splice) and
    the lower log loss on 1 (diabetes).}
  \label{tab:nano-per-dataset}
  \small
  \begin{tabular}{lrrrr}
    \toprule
    & \multicolumn{2}{c}{ROC AUC $\uparrow$} & \multicolumn{2}{c}{Log loss $\downarrow$} \\
    \cmidrule(lr){2-3}\cmidrule(lr){4-5}
    Dataset & WSP & SGD & WSP & SGD \\
    \midrule
    anneal                                     & 0.987 & 0.989 & 0.208 & 0.070 \\
    Bank\_Customer\_Churn                      & 0.849 & 0.854 & 0.399 & 0.348 \\
    blood-transfusion-service-center           & 0.694 & 0.717 & 0.501 & 0.501 \\
    churn                                      & 0.854 & 0.910 & 0.324 & 0.220 \\
    coil2000\_insurance\_policies              & 0.703 & 0.731 & 0.212 & 0.210 \\
    credit-g                                   & 0.771 & 0.778 & 0.532 & 0.505 \\
    diabetes                                   & 0.847 & 0.828 & 0.481 & 0.486 \\
    Fitness\_Club                              & 0.789 & 0.804 & 0.503 & 0.479 \\
    hazelnut-spread-contaminant-detection      & 0.889 & 0.949 & 0.545 & 0.291 \\
    Is-this-a-good-customer                    & 0.734 & 0.721 & 0.327 & 0.326 \\
    Marketing\_Campaign                        & 0.832 & 0.919 & 0.317 & 0.255 \\
    maternal\_health\_risk                     & 0.851 & 0.904 & 0.743 & 0.574 \\
    NATICUSdroid                               & 0.977 & 0.982 & 0.303 & 0.177 \\
    polish\_companies\_bankruptcy              & 0.873 & 0.944 & 0.205 & 0.124 \\
    qsar-biodeg                                & 0.886 & 0.915 & 0.399 & 0.338 \\
    seismic-bumps                              & 0.776 & 0.791 & 0.212 & 0.208 \\
    splice                                     & 0.995 & 0.991 & 0.430 & 0.153 \\
    students\_dropout\_and\_academic\_success  & 0.832 & 0.871 & 0.700 & 0.594 \\
    taiwanese\_bankruptcy\_prediction          & 0.924 & 0.940 & 0.136 & 0.084 \\
    website\_phishing                          & 0.915 & 0.974 & 0.461 & 0.303 \\
    \midrule
    Mean                                       & 0.849 & 0.876 & 0.397 & 0.312 \\
    \bottomrule
  \end{tabular}
\end{table}

\paragraph{Construction of WSP-NanoTabPFN.}
\label{app:nano-construction}
Summarised from Astra's final description. Each working scalar $s$ occupies two coordinates as $(s,-s)$, so a
cell's mean stays near zero; coordinates 190 and 191 hold $\pm65{,}536$ and dominate its variance, and every
LayerNorm, with gain $6688.74$ on the working coordinates, is then close to the identity on them; each norm thus acts as a fixed rescaling.
\begin{itemize}
  \item \textbf{Block 0} builds the representation: feature attention averages the target-cell indicator into
    $1/(f+1)$; the MLP bounds the standardised values to about $[-5,5]$, squares them (161 spline knots), turns
    training labels into class indicators and computes the gain $g(f)=0.9f^{0.65}$.
  \item \textbf{Block 1} estimates per-feature class probabilities: feature attention broadcasts the row's class
    indicators to every cell; five datapoint-attention heads smooth them within each column with Gaussian kernels
    of bandwidths 0.08, 0.20, 0.42, 0.90 and 1.80 (mixture weights 0.10, 0.25, 0.40, 0.20, 0.05), giving
    $p_{jc}(x)$, and the sixth averages them into class priors $\pi_c$; the MLP forms the evidence
    $e_{jc}=\log u_{jc}-\log(\pi_c+0.001)$ with $u_{jc}=0.97p_{jc}+0.03\pi_c+0.001$.
  \item \textbf{Block 2} measures each feature's reliability: datapoint attention averages the training rows'
    true-class log probability per column, which gives weights $w_j\propto\exp\big(2\,\tfrac{1}{n}\sum_i\log
    u_{j,y_i}(x_i)\big)$; the MLP scales the evidence by $g(f)$.
  \item \textbf{Block 3} combines the features: feature attention pools the evidence into row scores
    $S_c=g(f)\sum_j w_j e_{jc}$, and datapoint attention, with the training class indicators as keys and values,
    normalises them into $P_c\propto\pi_c\exp S_c$.
  \item \textbf{Block 4} builds a calibration correction: datapoint attention reuses the five kernels to average,
    within each feature neighbourhood, both the observed class indicators and the probabilities $P_c$ of the training
    rows; the MLP forms the log ratio of the two.
  \item \textbf{Block 5} applies it: feature attention adds the reliability-weighted correction to the row scores.
  \item \textbf{Decoder} reads out the smoothed log priors plus the scores, using
    $\mathrm{GELU}(t)-\mathrm{GELU}(-t)=t$ for a linear readout.
\end{itemize}
Logarithms use 25 knots over $[0.001,1.001]$, multiplications GELU's quadratic expansion at scale 0.003, and
selections attention logits of 40. Astra set every constant analytically, without trying it on data. It names
feature interactions as the main weakness, since a relation such as XOR is invisible to every per-feature estimate.

\paragraph{Prompt.}
\Cref{lst:nano-prompt} is the first message of the session, verbatim.

\begin{lstlisting}[caption={The prompt.}, label={lst:nano-prompt}, basicstyle=\ttfamily\scriptsize,
  columns=fullflexible, keepspaces=true, breaklines=true, frame=single]
TASK. Write, from scratch, the weights of a tabular foundation model: a transformer that classifies the test rows of a table in context, from the table's labelled training rows, so that its predictions on real classification datasets are as good as possible. All of its weights start at zero; you set them. The goal is the best possible predictions on the datasets described under HOW IT WILL BE MEASURED.

THE MODEL. The network is nanoTabPFN (the NanoTabPFNModel of TFM-Playground), a small reimplementation of the TabPFN v2 architecture. It is a prior-fitted network (PFN): such networks are normally trained by gradient descent on millions of synthetic tables drawn from a prior over datasets, so that, given a labelled training table and the test rows as one input, they output class probabilities for the test rows in a single forward pass, with no fitting at prediction time. Here it is not trained: you write its weights.

THE NETWORK. Input: one table with n_train labelled training rows, n_test test rows, f feature columns and C classes (2 <= C <= 10). Before the network, every table goes through this preprocessing, fitted on the training rows only (source: /work/source/interface.py, get_feature_preprocessor and TabularClassifier): constant columns are dropped; numeric columns are mean-imputed, with a 0/1 indicator column added for each column that has missing values; non-numeric columns are ordinal-encoded (categories numbered 0, 1, ... in sorted order; unseen categories become missing), mode-imputed and given an indicator in the same way; then every column is standardised (mean 0, standard deviation 1 over the training rows). Labels are encoded as 0..C-1 in sorted order.
The network (source: /work/source/nanotabpfn.py; embedding width E = 192, 6 attention heads of size 32, MLP width 768, 6 blocks; every linear layer computes y = x W^T + b with W stored as (out, in)):
  1. Cell embedding. Each feature column is standardised again over the training rows, z = clip((x - mean_train) / (std_train + 1e-8), -100, 100) (std with n - 1), and every cell becomes the 192-vector z * feature_encoder.linear_layer.weight[:, 0] + feature_encoder.linear_layer.bias. One target column is appended: training rows hold their label y in {0, ..., C-1} as a number, test rows hold the mean of y_train, and the target cell becomes y * target_encoder.linear_layer.weight[:, 0] + target_encoder.linear_layer.bias. The table is now a grid of (n_train + n_test) rows x (f + 1) cells of 192 numbers; the last cell of each row is its target cell. There are no positional or column embeddings.
  2. Six blocks, b = 0..5 (tensor prefix transformer_blocks.<b>.), each on the whole grid h:
     a. attention between features: within each row, all f + 1 cells attend to each other: h = norm1(h + MHA_features(h));
     b. attention between datapoints: within each column, every row (training and test) attends to the training rows of that column only (queries from all rows, keys and values from the training rows): h = norm2(h + MHA_datapoints(h));
     c. MLP on every cell: h = norm3(h + linear2(gelu(linear1(h)))), with the exact (erf) gelu.
     MHA is torch.nn.MultiheadAttention (self_attention_between_features, self_attention_between_datapoints): in_proj_weight (576 x 192) stacks W_q (rows 0-191), W_k (rows 192-383) and W_v (rows 384-575), and in_proj_bias stacks b_q, b_k, b_v; q = x W_q^T + b_q (likewise k, v); q, k and v are split into 6 heads of 32 consecutive dimensions; each head outputs softmax(q k^T / sqrt(32)) v; the heads are concatenated and passed through out_proj (weight 192 x 192, bias 192). Every norm is a LayerNorm over the 192 numbers of a cell: (h - mean) / sqrt(var + 1e-5) * weight + bias.
  3. Decoder: the target cell of each test row after block 5 gives 10 logits = decoder.linear2(gelu(decoder.linear1(h))). The predicted probabilities are the softmax over the first C logits.
Tensors: feature_encoder.linear_layer.weight (192 x 1), feature_encoder.linear_layer.bias (192), target_encoder.linear_layer.weight (192 x 1), target_encoder.linear_layer.bias (192); for each block b = 0..5: transformer_blocks.<b>.self_attention_between_datapoints.in_proj_weight (576 x 192), transformer_blocks.<b>.self_attention_between_datapoints.in_proj_bias (576), transformer_blocks.<b>.self_attention_between_datapoints.out_proj.weight (192 x 192), transformer_blocks.<b>.self_attention_between_datapoints.out_proj.bias (192), transformer_blocks.<b>.self_attention_between_features.in_proj_weight (576 x 192), transformer_blocks.<b>.self_attention_between_features.in_proj_bias (576), transformer_blocks.<b>.self_attention_between_features.out_proj.weight (192 x 192), transformer_blocks.<b>.self_attention_between_features.out_proj.bias (192), transformer_blocks.<b>.linear1.weight (768 x 192), transformer_blocks.<b>.linear1.bias (768), transformer_blocks.<b>.linear2.weight (192 x 768), transformer_blocks.<b>.linear2.bias (192), transformer_blocks.<b>.norm1.weight (192), transformer_blocks.<b>.norm1.bias (192), transformer_blocks.<b>.norm2.weight (192), transformer_blocks.<b>.norm2.bias (192), transformer_blocks.<b>.norm3.weight (192), transformer_blocks.<b>.norm3.bias (192); decoder.linear1.weight (768 x 192), decoder.linear1.bias (768), decoder.linear2.weight (10 x 768), decoder.linear2.bias (10). 116 tensors, 3,717,514 numbers. Only the values can be set; the architecture and the preprocessing are fixed.

HOW IT WILL BE MEASURED. After you finish, your network is evaluated on 20 classification datasets of the TabArena benchmark (TabArena-v0.1, curated real-world tabular datasets): those with at most 10,000 rows, 100 features and 10 classes. They have 748 to 10,000 rows and 4 to 85 features, numeric and categorical, some with missing values; 15 are binary and 5 have 3 to 5 classes. On each dataset, TabArena's first train/test split is used (two thirds training): the whole training part, up to 6,666 rows, is the context, and the network predicts every test row. Score: the one-vs-rest ROC AUC on each dataset, averaged over the 20 datasets (the main measure); the mean log loss (probabilities clipped at 1e-15) is reported too. You never see these datasets.

WHAT YOU HAVE.
1. A Python sandbox (no internet, no GPU) with a working folder /work that is kept between runs; each run is a fresh Python process, so only files in /work carry over. Available: numpy, scipy, scikit-learn, PyTorch (CPU), pandas, sympy, networkx. Before your code runs, these are defined: SHAPES and NAMES (the tensors), load(name) (a float32 numpy array: the values you saved, zeros at the start), save(name, array) (stores a tensor of the same shape) and saved() (the names saved so far). /work/source/ holds the source of the network and of the preprocessing. A run may take up to 900 seconds and 64 GB; its printed output (cut at 20,000 characters) is sent back to you. To run code, reply with exactly one block that starts with ```python. You may run code up to 120 times.
2. No pass tools (budget 0): your network cannot be run before the evaluation. Work from your design alone.
At the end, the tensors you saved with save() are your network; tensors never saved stay zero.

EXPLAIN EVERY STEP. In every reply, before your code block, explain in plain English (about 3 to 8 sentences): what the last output showed, what part of the network you will write or check now, what it should compute and why that serves the goal, and how you will know from the output that it was written as intended. These explanations are read by the experimenters.

RULES. Not allowed: (a) gradient descent or any other training: no backward passes, autograd or optimizers, and no fitting of weights to data or labels (no regression, least squares or other fitting on examples, whether practice data or tables you generate yourself); (b) running the network outside the pass tools: do not implement or simulate the network's computation, or any part of it, on any input in the sandbox; (c) internet access (there is none). Allowed: writing the weights by code from your own design (formulas, loops, symmetries, linear algebra on the weights themselves).

HOW TO FINISH. When your network is saved, reply WITHOUT a code block and describe your network in detail, part by part: the overall idea (what classifier it implements and why you chose it); how the 192 dimensions of a cell are laid out and why the norms preserve them; what the feature attention, datapoint attention and MLP of each block compute, and the decoder; the main numerical choices (scales, temperatures, kernel widths, priors) and how you set them; and where you expect it to be weakest.
\end{lstlisting}

\clearpage
\section{Signed-product intervention: setup and analysis}
\label{app:signed-products}

This section documents the task supplied to the agent, its operating conditions, and the evaluation of the
weight intervention in \cref{sec:exp-failure}.

\paragraph{Experimental setup.}
We investigate whether an LLM agent can diagnose a specific limitation in a pretrained TFM and modify its weights
to improve the corresponding computation. We consider noiseless signed-product regression,
\[
  y = \prod_{i=1}^{k} x_i, \qquad x_i = s_i \exp(u_i),
\]
where signs $s_i \in \{-1,+1\}$ are sampled independently with equal probability and
$u_i \sim \mathcal{U}(-0.5,0.5)$. We evaluate $k \in \{5,6\}$ inputs and $N \in \{256,512\}$ context rows
using the official TabPFN-3.5 checkpoint~\citep{tabpfn35} with the default regressor configuration.

We give Astra access to the model source, pretrained weights, 12 labelled training datasets per regime, and a
suite of 15 OpenML datasets for checking preservation locally. The agent may inspect activations, run forward
passes, and perform direct weight edits and closed-form linear solves; gradient-based optimisation is prohibited.
It submits three candidates for validation feedback before selecting one final edit. We evaluate that edit on
60 hidden datasets per regime, each containing 500 query rows, and assess preservation on 51 TabArena-Lite tasks.
The architecture, preprocessing, and inference procedure remain unchanged.

\paragraph{Model and submitted artifact.}
We use \texttt{TabPFNRegressor} from TabPFN 9.2.0 with the official checkpoint
\path{tabpfn-v3.5-20260909.safetensors} on CUDA, in its default configuration (eight ensemble members) with
\texttt{random\_state=0}. The agent submits one \texttt{safetensors} file of replacement tensors with the original
names, shapes, and dtypes. The evaluator merges them into the pretrained checkpoint and runs the ordinary estimator;
the regression output borders are not editable, and no agent-written code runs at evaluation.

\paragraph{Agent and interaction.}
One GPT-6 Astra agent (\texttt{gpt-6-astra}) runs through the Codex CLI at reasoning effort \texttt{xhigh}. Unlike
in the nanoTabPFN setup, it keeps its shell tools, so it can write scripts, inspect activations, and run candidate
models; no human answers questions during the run. Its first message concatenates the rules, the task prompt, and a
note with the workspace paths (\cref{lst:product-rules,lst:product-prompt,lst:product-interaction}). The agent
validates a candidate file with \texttt{kvp wcheck}, requests validation feedback with \texttt{kvp feedback}, and
records its single final edit with \texttt{kvp submit}. Feedback exposes only aggregate validation statistics and
preservation outcomes, including the names of tasks over their error limits. Requests are not capped, but waiting
for them counts toward the five-hour wall-clock budget, and none is available after submission. The agent records
its hypotheses, predictions, and controls in \texttt{notes.md}.

\paragraph{Sandbox.}
The CLI runs in fresh Linux user, mount, IPC, and PID namespaces with a restricted filesystem root. The checkpoint,
the Python environment, and the model and harness sources are mounted read-only; the workspace is writable, and one
GPU is exposed. The evaluator, the experiment configuration, other runs, and the hidden validation and test data
are not mounted, and feedback is scored outside the sandbox. The network stays open for the remote agent, so the
source restrictions below are checked by audit, not enforced by isolation.

\paragraph{Rules and audit.}
The rules prohibit computed gradients (including finite differences and hand-derived updates), iterative loss
minimisation, and automated search guided by a loss. They permit closed-form solves (least squares, ridge, SVD),
activation analysis, and comparison of individually reasoned edits. A Python startup patch disables autograd and
PyTorch's optimisers. External benchmark data and results are off limits. A causal claim requires a prediction
written in \texttt{notes.md} before the intervention and a corresponding control.

Our audit covered both CLI turns, all 143 commands, and all 35 agent-written scripts. It found no gradient
computation, iterative weight optimisation, prohibited source access, or attempt to bypass the restrictions; every
fitted map is a closed-form ridge solve. The source-access conclusions rest on the recorded commands and events,
not on network packet captures.

\paragraph{Hardware and software.}
The session ran on 11~November~2026; Astra itself was served remotely. The SLURM launcher requests three NVIDIA
L40S GPUs (one for the agent, two for feedback scoring), 16 CPU cores, and 160\,GB of memory. The project requires
Python~3.12 or newer, and its lockfile pins TabPFN~9.2.0, PyTorch~2.15.0, NumPy~2.5.4, SciPy~1.18.1, scikit-learn~1.9.1, and
safetensors~0.8.0.

\paragraph{Evaluation protocol.}
For each of the four $(k,N)$ regimes, the harness provides 12 labelled training datasets, uses 20 separate validation
datasets for feedback, and reserves 60 test datasets for the final evaluation. Every dataset contains $N$ context rows and 500 query
rows. The training datasets include query labels, so the agent can analyse and fit readouts on these examples. Each
dataset is drawn from its own random stream, determined by a secret seed, its split (training, validation, or test),
its regime, and its index; the agent never sees the seed, so the three splits do not overlap. Test results are
computed only after the final submission.

The OpenML suite contains five regression datasets (kin8nm, cpu\_act, abalone, space\_ga, wind), six binary
ones (phoneme, banknote-authentication, ozone-level-8hr, spambase, ilpd, wdbc), and four multiclass ones (vehicle,
segment, cmc, satimage), each with at most 2,000 context rows. None shares an OpenML identifier or name with the
51 TabArena-Lite tasks, and known near-duplicates of TabArena tasks were excluded by hand. Besides checking
preservation, Astra used activations from these datasets to fit the imputation maps and the gate; since the suite is
disjoint from TabArena-Lite, this does not affect the final preservation score.

We compute $R^2$ and normalised mean squared error (nMSE: squared error divided by target variance) over all query
rows. Sign accuracy is measured on queries whose target magnitude is at or above that dataset's 25th percentile.
The success criteria, fixed before the run, require each regime to satisfy: median $R^2\geq0.90$, at least 90\% of datasets with
$R^2\geq0.80$, summed edited nMSE at most half the pretrained model's, at least 90\% of datasets with sign accuracy
at least 0.95, and no invalid predictions. These criteria are fixed throughout the session.

Preservation is evaluated on 30 binary-classification, eight multiclass-classification, and 13 regression tasks.
Per-task errors are $1-\mathrm{AUROC}$, log loss, and RMSE, respectively, averaged over seeds 0, 1, and 2 on the
official outer test splits. We then form edited-to-pretrained error ratios using fixed small error floors and take
their geometric mean. The preservation criterion requires this mean to be at most 1.01, every task to remain within
its preset error-ratio limit, and no invalid predictions. Feedback instead uses inner splits of the official
training rows with seed 0. The reported 95\% interval is a percentile bootstrap over tasks with 10,000 resamples.

\paragraph{Session and feedback.}
The agent requested validation feedback on three candidates (\cref{tab:product-session}). The first failed one
product regime and the preservation check, the second passed the product criteria but exceeded one task's
preservation limit, and the third, \texttt{count\_gated\_v5}, passed both and was submitted.
\begin{table}[ht]
  \centering
  \caption{The signed-product intervention session. Candidate counts refer to requests for evaluator feedback,
    not every local diagnostic or edit.}
  \label{tab:product-session}
  \small
  \begin{tabular}{lr}
    \toprule
    Agent sessions & 1 \\
    CLI turns (including final notes) & 2 \\
    Wall-clock time / allowed budget & 62 min / 5 h \\
    Recorded commands / scripts & 143 / 35 \\
    Feedback requests / final submissions & 3 / 1 \\
    Changed tensors / weight entries & 125 / approximately 105 million \\
    \bottomrule
  \end{tabular}
\end{table}

\begin{figure}[ht]
  \centering
  \includegraphics[width=\linewidth]{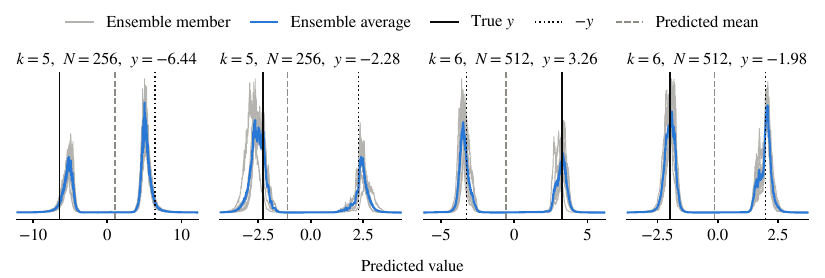}
  \caption{Predictive distributions of pretrained TabPFN-3.5 for four signed-product queries, from two freshly
    generated datasets. Each of the eight ensemble members (grey) places separate modes near $y$ and $-y$, and so
    does their average (blue), which the estimator uses; its mean (dashed) therefore lies near zero.}
  \label{fig:product-bimodal}
\end{figure}

\paragraph{Diagnosing the limitation.}
Pretrained TabPFN-3.5 predicts values near zero on the signed-product tasks (median $R^2$ between $-0.024$ and
$0.017$), yet accurately predicts positive-only products with the same input magnitudes. Its predictive distribution is
bimodal, with modes near $+|y|$ and $-|y|$ that cancel in the mean (\cref{fig:product-bimodal}). As an oracle
diagnostic, Astra moves this mass to the side of the known target sign, which raises $R^2$ to $0.993$--$0.999$ on one
exploratory dataset per regime. The magnitude is therefore already available, and selecting the sign is the main
obstacle; the final intervention must infer the sign from the model's internal representations.

We confirmed this independently on 20 freshly generated datasets, five per regime, drawn with the task's generator
and our own seed. On every dataset, a median of 82--96\% of each query's predictive mass lies within 25\% of $+|y|$
or $-|y|$, and the median probability of a negative value is 0.46--0.56, so the distribution says nothing about the
sign. The two modes are not a disagreement between ensemble members: in 88--100\% of member--query pairs, the member
itself places between 20\% and 80\% of its mass on negative values. Folding the mass onto the true sign raises $R^2$
from $-0.18$--$0.14$ to $0.993$--$0.999$, reproducing the agent's oracle.

\paragraph{Mechanistic intervention.}
We identify sign-relevant information in the pretrained model and construct a weight intervention that makes this
information available to its prediction head. In the agent's recorded analysis, linear probes recover the number of
negative inputs from early representations, even where directly predicting the product's sign with a linear probe
performs poorly. The distinction matters because the product is negative when this count is odd and positive when
it is even: the required sign is a nonlinear function of the count.

Building on these observations, the intervention writes a count readout into reserved residual-stream coordinates,
preserves it through subsequent layers, and constructs a parity computation in the regression head that favours
the corresponding output sign. Additional closed-form weight adjustments compensate for repurposed coordinates,
while an internal gate limits the intervention on other tasks. In the agent's controls, reserving the same
coordinates without activating the sign output leaves performance near chance. These results support the role of
the constructed sign computation in the improvement; they do not establish that the pretrained model already
contained the complete circuit.

\paragraph{Results and scope.}
On the hidden test datasets, the edited model raises median $R^2$ from approximately zero to $0.991$--$0.998$
(\cref{tab:signed-product-results}). Between 95\% and 98\% of datasets in each regime achieve $R^2 \geq 0.8$.
On TabArena-Lite, the geometric-mean edited-to-pretrained error ratio is $1.0008$
(95\% confidence interval: $0.9999$--$1.0017$), indicating little aggregate change on the preservation benchmark.
The task-type ratios in \cref{tab:product-preservation} show slightly lower regression error and slightly higher
classification error; these small differences are descriptive. With a $\pm0.5\%$ tie band, the edited model wins
on one task, ties on 47, and loses on three. No task exceeds its preservation limit or produces invalid predictions.
All four signed-product regimes pass the fixed success criteria. Median sign accuracy is 1.00 after the edit,
compared with 0.50--0.56 before it. Loading the merged checkpoint directly into the ordinary estimator reproduces
the reported scores on all 240 hidden datasets, confirming that inference requires only the modified weights.
Transfer beyond the targeted regimes is partial. The intervention also changes approximately 105 million weight
entries, so it does not establish that a small local edit suffices. Together, these results demonstrate that
agent-guided analysis and weight programming can make sign-relevant information useful for prediction while
largely preserving broader benchmark performance.

\begin{table}[ht]
  \centering
  \caption{Signed-product prediction on 60 hidden datasets per regime, each with $k$ inputs, $N$ context rows,
    and 500 query rows. The last columns report the edited model's share of datasets with $R^2\geq0.8$ and
    the ratio of summed edited to pretrained nMSE. These are intervention results, not oracle diagnostics.}
  \label{tab:signed-product-results}
  \small
  \begin{tabular}{rrrrrr}
    \toprule
    $k$ & $N$ & \shortstack{Pretrained\\median $R^2$} & \shortstack{Edited\\median $R^2$}
      & \shortstack{Share with\\$R^2\geq0.8$} & \shortstack{nMSE\\ratio} \\
    \midrule
    5 & 256 & $-0.018$ & $0.995$ & 96.7\% & 0.022 \\
    5 & 512 & $0.017$  & $0.998$ & 98.3\% & 0.007 \\
    6 & 256 & $-0.024$ & $0.991$ & 95.0\% & 0.036 \\
    6 & 512 & $-0.024$ & $0.997$ & 98.3\% & 0.009 \\
    \bottomrule
  \end{tabular}
\end{table}

\begin{table}[ht]
  \centering
  \caption{Preservation on TabArena-Lite. Ratios below one indicate lower error after the intervention.
    Per-task errors are averaged over three seeds before aggregation.}
  \label{tab:product-preservation}
  \small
  \begin{tabular}{lrr}
    \toprule
    Task type & Tasks & Geometric-mean error ratio \\
    \midrule
    Binary classification & 30 & 1.0009 \\
    Multiclass classification & 8 & 1.0020 \\
    Regression & 13 & 0.9997 \\
    \midrule
    All tasks & 51 & 1.0008 \\
    \bottomrule
  \end{tabular}
\end{table}

\paragraph{Prompt and rule materials.}
\Cref{lst:product-rules,lst:product-prompt} reproduce the rules and the task prompt as excerpts; omission markers
identify removed passages and repository names, and retained passages are unchanged. The prompt calls the training
datasets ``public'' datasets and a regime a ``stratum''.
\Cref{lst:product-interaction} gives the driver's message templates, with runtime values replaced by placeholders.
Baseline values quoted in the prompt belong to the problem specification, not to the hidden-test results in
\cref{tab:signed-product-results}.

\begin{lstlisting}[caption={Weight-editing rules supplied to Astra (excerpt).}, label={lst:product-rules},
  basicstyle=\ttfamily\scriptsize, columns=fullflexible, keepspaces=true, breaklines=true, frame=single]
# Your task: repair TabPFN-3.5 by editing its weights

TabPFN-3.5 is a transformer that is fitted on a dataset's context rows and predicts at query rows in one forward
pass, without training. On some kinds of data it fails. Your experiment's `PROMPT.md` (below) states the failure, how
it is measured and the bars a repair must reach. Your job is to understand why the model fails, by reasoning about
its internals, and to fix it by editing the values of its weights, while keeping what the model does well elsewhere.

The TabPFN source calls the context rows *train* rows and the query rows *test* rows.

## What you hand in

One weights file for every dataset: a safetensors file holding the tensors you changed (any subset of the
checkpoint's tensors), each with its stock name, shape and dtype, all values finite. The evaluator loads the stock
checkpoint, replaces those tensors with yours and runs the ordinary estimator. The checkpoint's other settings stay as
they are.

```python
from kvp.weights import check, merged, regressor, classifier, changed, sha256
check("candidates/a.safetensors")              # raises ValueError with the reason if the file is not acceptable
path = merged("candidates/a.safetensors", "work/merged")   # a full checkpoint with your tensors in place
reg = regressor(path)                          # the stock estimator configuration, with your weights
clf = classifier(path)
```

`regressor(checkpoint)` with the stock checkpoint path gives the stock model. Run Python with `python` (TabPFN, torch,
numpy, scipy and scikit-learn are installed).

## What you may change

- **Editable:** the learned tensors of the checkpoint, except `heads.regression_borders` (the regression output
  borders), which stays fixed.
- **Fixed:** the architecture, preprocessing, ensembling and inference settings, and the estimator configuration. No
  wrappers, hooks, extra code paths, or switching on properties of the dataset: the evaluator runs the stock code with
  your tensors, nothing else of yours.

## The rules

**1. No computed gradients and no training**, in everything you run. Do not compute gradients of any kind: not by
autograd, not by finite differences (changing a value slightly to see how a loss moves), not by hand-derived
gradient formulas. Do not train: no procedure that updates values step by step to reduce a loss (gradient descent,
torch optimisers, `scipy.optimize`, iterative solvers), and no evolutionary, random or other automated search guided
by a loss.

Allowed: closed-form computations (least squares, ridge, pseudoinverse, SVD, eigendecompositions, projections) and
other linear algebra and statistics; reading and analysing any tensor or activation; trying several candidates that
you reasoned out by hand and keeping the best of them. Python here starts with autograd and torch optimisers switched
off; getting around that, or installing other libraries to do it, breaks the rule. Every command you run is recorded
and audited.

**2. Sources.** You have the network. Do not open TabArena or OpenML datasets, task pages, results or leaderboards, and
do not open the repositories [repository names omitted], local
copies or on GitHub. Papers, documentation and the TabPFN source are fine.

**3. Evidence.** A claim that a component causes the failure needs a prediction written in `notes.md` before the
intervention, the targeted edit, and a control (undoing the edit, or a random edit of the same size); without that,
call it a correlation.

A run that breaks rule 1 or 2 does not count, whatever its score.

## Commands

- `kvp wcheck <file>`: validates a candidate here, as the evaluator will, and shows which tensors and how many entries
  it changes. Decides nothing.
- `kvp feedback <file>`: sends a copy of the candidate to the evaluator for scored feedback (`PROMPT.md` says what it
  returns). Feedback is queued and scored one at a time; the result appears in `inbox/<id>.json`. Unlimited, but queue
  and scoring time count against your time limit. Passing feedback means "ready for final scoring", not success:
  the final score uses other, hidden data.
- `kvp results [<id>]`: the feedback returned so far.
- `kvp status`: your deadline, the time left, the feedback queue, and your final submission.
- `kvp submit <file>`: your final submission. Its hash is recorded at once and it is the one file scored at the end.
  Only one; the run closes when your turn ends. Submitting nothing is allowed.

## Time and conduct

- You have **5 hours** of wall-clock time from the start of the run, including every wait. `kvp status` shows the
  deadline; it can only move earlier, and you are told if it does. At the deadline the run stops, also in the middle of
  a command; the last final submission stands, or none.
- Nobody answers questions during the run: decide yourself and continue.
- Keep `notes.md` in your workspace: each hypothesis, the evidence for and against it, the explanations you rejected,
  and what you submitted and why.
- "No qualifying repair found" is an acceptable outcome. Report it as such rather than submitting something you do
  not believe in.
\end{lstlisting}

\begin{lstlisting}[caption={Signed-product task prompt supplied to Astra (excerpt).}, label={lst:product-prompt},
  basicstyle=\ttfamily\scriptsize, columns=fullflexible, keepspaces=true, breaklines=true, frame=single]
# Experiment: products of signed inputs

## The failure

Stock TabPFN-3.5 (the official checkpoint `tabpfn-v3.5-20260909.safetensors`, `TabPFNRegressor` with every argument at
its package default, CUDA) fails to learn y = x_1 x_2 ... x_k from its context rows once there are 5 or more inputs of
both signs, although those rows determine the answer: a fit that knows the product form (the sign of the input product
times exp of a least-squares fit of log|y| on log|x_i|, using only the context rows) recovers every such dataset exactly.

A dataset here is one product rule with k inputs: x_i = s_i exp(u_i), s_i = +1 or -1 with equal chance, u_i ~ U(-0.5,
0.5), y = x_1 ... x_k, noiseless. N context rows, 500 query rows from the same law.

[... additional baseline comparisons omitted ...]

## The regime you must repair

- The law above, strata (k, N) in {(5, 256), (5, 512), (6, 256), (6, 512)}, 500 query rows, all queries scored.
- **Per dataset:** R^2 on the queries; nMSE = mean squared error / variance of the query y; sign accuracy on queries
  with |y| at or above the dataset's 25% quantile of |y|.
- **Stock on this regime** (fresh datasets, 60 per stratum): median R^2 -0.01 / 0.03 / -0.03 / -0.01, share >= 0.8 0
  in every stratum, median sign accuracy 0.51-0.57.

## Bars (frozen; every stratum must pass)

- median R^2 >= 0.90;
- at least 90% of datasets with R^2 >= 0.80;
- sum of nMSE (your model) / sum of nMSE (stock) over the same datasets <= 0.50;
- at least 90% of datasets with sign accuracy >= 0.95;
- no invalid prediction (NaN, crash).

**And preservation:** the same weights must keep TabPFN-3.5's general performance. The final submission is scored on
the hidden TabArena-Lite benchmark (51 real datasets: 30 binary, 8 multiclass, 13 regression; official test rows;
errors 1-AUROC, log loss, RMSE; per-task error averaged over seeds 0, 1, 2). The geometric mean over tasks of
(your error / stock error), with small per-type floors, must be at most **1.01**, no single task may get worse than its
per-task limit (1.10 for most tasks, up to 1.50 for the noisiest), and no task may produce invalid predictions.
Classification uses the same checkpoint, so edits to shared parts of the model affect it.

[... supplementary evaluation instructions omitted ...]

## Your data and tools

- `data/family/<stratum>_<i>.csv`: 12 public datasets per stratum (`role` = context or query; query `y` is given). The
  hidden validation and test datasets come from the same generator with other seeds.
- `data/devsuite/` and `data/devsuite_score.py`: 15 real OpenML datasets (none in TabArena) to check preservation
  yourself (coarse: its own seed noise is about 1%). See `data/README.md`.
- A GPU is available to you (`device="cuda"`).
- **Feedback** (`kvp feedback <file>`) returns, for your candidate: per stratum on 20 hidden validation datasets
  (median R^2, share >= 0.8, nMSE ratio, sign statistics, pass), and TabArena-Lite on inner splits of the official
  training rows (geometric-mean ratio overall and per type with a 95% interval, tasks over their limit by name, invalid
  tasks, pass), plus `ready_for_final`. It takes several minutes per candidate; it counts against your time.
\end{lstlisting}

\begin{lstlisting}[caption={Driver interaction templates for the signed-product session.}, label={lst:product-interaction},
  basicstyle=\ttfamily\scriptsize, columns=fullflexible, keepspaces=true, breaklines=true, frame=single]
# Driver messages (templates)

The first message concatenates RULES_WEIGHTS.md, PROMPT.md, and this paths note:

Your workspace (the current folder): `<workspace>`. The harness code: `<harness-source>`. The TabPFN source: `<tabpfn-source>`. The stock checkpoint (read only): `<checkpoint>`. You have one GPU (`cuda`).

If a turn ends while the run is open:

The run is still open, and nobody answers questions. Continue. Time left: <remaining minutes and deadline>. `kvp status` shows the deadline and your feedback queue, `kvp results` the returned feedback. When you have decided, run `kvp submit <file>.safetensors` (or submit nothing).

After the final submission:

Your final submission is recorded (sha256 <candidate hash>). The run is closed: no more feedback. Write the final section of notes.md: what you submitted and why, what you expect its score to be, and what you learned. Then stop.
\end{lstlisting}

\clearpage
\section{Programming to steer fitted function: setup and programs}
\label{app:kvp}

This section gives the setup and submitted programs for \cref{sec:exp-kv}.

\paragraph{Model and cache.}
We use \texttt{TabPFNRegressor} from TabPFN~9.2.0 with the official
checkpoint \path{tabpfn-v3.5-20260909.safetensors}, one estimator,
\texttt{fit\_mode="fit\_with\_cache"}, and float32 inference on CPU.
The KV cache is a flat dictionary of tensors: context keys and values for
24 layers, inducing hidden states, scaler means and standard deviations,
and \texttt{ecdf\_context}, which stores sorted context inputs without their
assignment to cache rows.
An edit must return every tensor with its original name, shape and dtype
and finite entries. Only keys, values and inducing hidden states may change.
At test, the program receives only this cache. It may call the black-box
\texttt{recompute(cache, layer)}, which recomputes the caches above the
specified layer using the model held by the harness.
Neither submitted program calls it.
All layer indices below start at zero.

\paragraph{Agent and interaction.}
The agent is GPT-6 Astra (\texttt{gpt-6-astra}) through the Codex CLI,
with reasoning effort \texttt{xhigh} and reasoning summaries \texttt{detailed}.
Each turn uses \texttt{codex exec} initially or
\texttt{codex exec resume} for the existing session.
The external jail described below provides isolation.
The agent keeps its shell tools and system prompt.
During development, it has the model weights, TabPFN source, harness code
and training datasets, and can write scripts and run forward passes freely.
No human answers questions during the run.
The first message concatenates \texttt{RULES.md}, the task's
\texttt{PROMPT.md}, and a note giving the workspace, harness and TabPFN source
paths. The agent records what it tries and learns in \texttt{notes.md}.
It runs \texttt{kvp check edits/<name>.py} to execute a candidate on the
training datasets in the edit sandbox and receive its printed output and
prediction errors. This is a development check.
It submits with \texttt{kvp submit edits/<name>.py}.
After a turn without a submission, the driver resumes the session with an
instruction to continue.
After a turn with a submission, the driver runs \texttt{kvp test} outside
the jail, evaluates the submitted file once on the hidden test datasets,
and closes the run. It then resumes the session once with the result and
an instruction to write final notes.
Reaching the six-hour limit without submission closes the run without a test.
The driver finishes by running \texttt{kvp audit}.

\paragraph{Sandbox.}
The agent CLI runs in fresh Linux user, mount, IPC and PID namespaces with
a restricted filesystem root.
The weights, Python environment, TabPFN source and harness source are
mounted read-only; its workspace and private home are writable.
The network remains available for the remote agent.
The experiment configuration, other runs and test datasets are not mounted.
Each edit runs in a separate process with an additional network namespace,
without weights, experiment files or network access.
The model stays in the harness process; calls to \texttt{recompute} are
served over a pipe.
The edit runs under PyTorch inference mode and has a 900\,s limit per dataset.

\paragraph{Rules and audit.}
Both tasks retain all three rules: no computed gradients or training,
no learned cache maps, and no precomputed artifacts, with programs capped
at 32,768 bytes. The full rules appear in \cref{lst:kvp-rules}.
Python in the agent jail starts with autograd and PyTorch optimisers disabled.
Static checks reject prohibited edit imports and names.
The harness checks the returned cache for the required tensor names, shapes,
dtypes and finite entries, and for changes to protected tensors.
An automatic audit scans recorded commands and code for possible violations.
We read the transcripts to check for learned cache maps and read both the
transcripts and submitted programs to check for precomputed artifacts.

\paragraph{Hardware and software.}
The runs took place on 11~November~2026.
Model fitting and prediction use CPU; Astra itself is served remotely.
The project selects Python~3.12 and requires Python~3.12 or newer.
Its lockfile pins TabPFN~9.2.0, PyTorch~2.15.0, NumPy~2.5.4,
SciPy~1.18.1, scikit-learn~1.9.1 and safetensors~0.8.0.

\paragraph{Datasets and evaluation.}
Each dataset has one input feature.
Its interval has centre $m\sim\mathcal{U}[-2,2]$ and width
$L\sim\mathcal{U}[2,6]$.
Context and query inputs are drawn uniformly from $[m-L/2,m+L/2]$.
For \emph{flip the slope}, labels follow
\[
  y=wx+b+\epsilon,\qquad
  |w|\sim\mathcal{U}[0.5,3],\quad
  b\sim\mathcal{U}[-1,1],\quad
  \epsilon\sim\mathcal{N}(0,0.05^2),
\]
where $w$ is the slope, $b$ is the intercept and $\epsilon$ is independent
observation noise. Either sign of $w$ is equally likely.
For \emph{sine to square wave},
\[
  y=b+A\sin(2\pi k u+\phi)+\epsilon,\qquad
  u=\frac{x-(m-L/2)}{L},
\]
where $u$ is position within the interval,
$A\sim\mathcal{U}[0.5,1.5]$ is the amplitude,
$k\sim\mathcal{U}[0.5,2]$ is the number of cycles across the interval,
and $\phi\sim\mathcal{U}[0,2\pi)$ is the phase.
The offset $b$ and noise $\epsilon$ have the distributions above.
Parameters are drawn independently for each dataset.
The context size is drawn uniformly from the integers 32--128; each dataset
has 500 query rows.
The development phase provides three training datasets for \emph{flip the slope}
and five for \emph{sine to square wave}, including all inputs and labels.
The test phase uses ten unseen datasets per task from the same family,
drawn independently of the development data.

Let $\hat y_j$ be the original TFM prediction at query input $x_j$, and let
$\bar x$ be the mean context input.
For \emph{flip the slope}, fit the line $\hat w x+\hat b$ by least squares
to $(x_j,\hat y_j)$, where $\hat w$ and $\hat b$ are its slope and intercept.
The target is
\[
  g_{\mathrm{flip}}(x)
  =(\hat w\bar x+\hat b)-\hat w(x-\bar x).
\]
For \emph{sine to square wave}, fit
$\hat c+\hat A\sin(\hat\omega x+\hat\phi)$ to the original predictions,
where $\hat c$, $\hat A$, $\hat\omega$ and $\hat\phi$ are the fitted
offset, amplitude, angular frequency and phase.
The target is
\[
  g_{\mathrm{square}}(x)
  =\hat c+\hat A\,\operatorname{sign}
       \bigl(\sin(\hat\omega x+\hat\phi)\bigr).
\]
The sine fit searches 0.1--4 cycles across the query-input span in steps of
0.002 cycles. At each frequency, its constant, sine and cosine coefficients
are fitted by linear least squares; the fit with the smallest squared error
is retained.
These targets, rather than transformations of the noisy labels, are the
dashed curves in the figures.

Let $\tilde y_j$ be the edited prediction and $M=500$ the query count.
For selecting the best, median and worst datasets in \cref{fig:kv-cache},
we rank each task's test datasets by the normalised root mean squared error,
\[
  \mathrm{nRMSE}
  =\frac{\sqrt{M^{-1}\sum_{j=1}^{M}(\tilde y_j-g(x_j))^2}}
  {\sqrt{M^{-1}\sum_{j=1}^{M}(\hat y_j-\bar{\hat y})^2}},
  \qquad
  \bar{\hat y}=M^{-1}\sum_{j=1}^{M}\hat y_j .
\]
Here $g$ is the corresponding target and $\bar{\hat y}$ is the mean original
prediction.
In this paper, we use this quantity only to rank the displayed examples.

\paragraph{Session.}
We ran one session per task under the six-hour limit.
The slope program was submitted after 19.6 minutes and the square-wave
program after 70.3 minutes.
Both automatic audit files contained no matches.
Transcript and submitted-program reviews found no rule violations.

\subsection{The submitted programs}
\label{app:kvp-edits}

Write $K_\ell$ and $V_\ell$ for the context-key and context-value matrices
at layer $\ell$, with one row per context example.
All coordinates, fits and matrices below are computed from the supplied cache.

\paragraph{The slope program.}
The program estimates position from the main direction of variation in early
values, refines its spacing with the sorted inputs, and reflects the smooth
part of later values. The residual of each polynomial fit is retained.
Its five named settings are shown in \cref{lst:kvp-slope}.

\begin{lstlisting}[caption={The slope program: settings and pseudocode.},
  label={lst:kvp-slope}, basicstyle=\ttfamily\scriptsize,
  columns=fullflexible, keepspaces=true, breaklines=true, frame=single]
COORDINATE_LAYER = 2       # Values used to estimate position.
FIRST_EDIT_LAYER = 18      # First value layer to change.
POLYNOMIAL_DEGREE = 5      # Degree of the smooth fits.
REFLECTION_GAIN = 1.3      # Strength of reflection.
COORDINATE_BLEND = 0.5     # Weight of sorted-input spacing.

For each cache batch:
  R = column-centred V_2
  z = standardised first left singular vector of R
  For every feature and both orientations:
    Recover ordered inputs from ecdf_context by interpolation.
    Assign these inputs to rows in the rank order of z.
    Standardise the resulting coordinate q.
    Fit each column of R as a degree-5 polynomial in q.
  Keep q with the smallest squared reconstruction error.
  Reverse q if needed to align its sign with z.
  z = standardise(0.5*z + 0.5*q)
  t = z / sqrt(1 + z*z)

  For layers l = 18,...,23 and every value coordinate:
    P = least-squares degree-5 fit of V_l against z
    Q = least-squares degree-5 fit of V_l against t
    reflected_z = clip(-z, min(z), max(z))
    V_l += (1.3/2) * (P(reflected_z) - P(z)
                     + Q(-t) - Q(t))
Return the full cache; keys and other tensors are unchanged.
\end{lstlisting}

The reflection in $z$ is clipped to avoid polynomial extrapolation beyond
the observed range; the reflection in $t$ is not clipped.
The steepest test line comes out too shallow and sags to the right of
the crossing.

\paragraph{The square-wave program.}
The program orders rows using key similarities, finds a sine supported by both
early-key geometry and late values, and moves the fitted part of each value
toward a peak or trough.
It retains the fitted residual and scales the layer-21 keys to concentrate
attention.
The submitted file has 16 named settings, listed in \cref{lst:kvp-square}.

The program reads a label coordinate from a conic traced by the layer-0 keys.
Candidate versions of this coordinate are used only to check candidate sines;
they do not supply target caches.

\begin{lstlisting}[caption={The square-wave program: settings and pseudocode.},
  label={lst:kvp-square}, basicstyle=\ttfamily\scriptsize,
  columns=fullflexible, keepspaces=true, breaklines=true, frame=single]
POSITION_LAYER = 17         # Keys used to order rows.
SIGNAL_LAYER = 21           # Values used to find the sine.
MAX_RANK = 8                # Largest value subspace.
NEIGHBORS = 5               # Neighbour for graph bandwidth.
MIN_CYCLES = 0.4            # Lower sine-search bound.
MAX_CYCLES = 2.1            # Upper sine-search bound.
CYCLE_STEP = 0.002          # Sine-search resolution.
MIN_CORRELATION = 0.85      # Agreement with first value direction.
SKEW_STEP = 0.01            # Label-coordinate search resolution.
GEOMETRY_TOLERANCE = 1.5    # Early-key error tolerance.
LABEL_MIX = 0.0             # No decoded-sign contribution.
KEY_SCALE = 1.5             # Layer-21 key multiplier.
BASE_GAIN = 2.0             # Layer-21 base transport gain.
EDGE_GAIN = 4.0             # Extra gain near sine crossings.
PREVIOUS_GAIN = 0.5         # Layer-20 gain.
FINAL_GAIN = 1.0            # Layer-23 gain.

Take the first three left singular vectors of uncentred K_0.
Fit a homogeneous quadratic form to their rows.
If the form has the required conic signature:
  Transform rows to angles using its eigenvectors and eigenvalues.
  Circularly centre the angles.
  q = tangent of each half-angle, scaled to max(abs(q)) = 1.
  Candidate label coordinates = q / (1 + alpha*q),
    centred and normalised, for alpha from -0.98 to below 0.99
    in steps of 0.01.
Otherwise mark the conic check unavailable.

D_ij = squared distance between rows i and j of K_17.
h = mean squared distance to each row's fifth neighbour.
W_ij = exp(-D_ij / h).
Order rows by the second-smallest-eigenvalue eigenvector
  of diag(W*1) - W, where 1 is the all-ones vector.
Rescale the sorted cached inputs to [0,1].
U = left singular vectors of column-centred V_21.

For rank r = 3,...,min(8, floor(number_of_rows/4), columns(U)):
  For both assignments of sorted inputs to the row order,
      and frequencies from 0.4 to 2.1 cycles in steps of 0.002:
    Centre the sine and cosine columns at this frequency.
    Find their unit-norm combination with least squared
      residual outside span(U[:, :r]); this fixes the phase.
    Record this residual and the absolute correlation
      with U[:, 0].
  Keep the smallest-residual candidate with correlation
    at least 0.85; if none qualifies, omit that constraint.
  Form its uncentred, unit-amplitude sine s.
  Geometry error = 1 - largest squared correlation of s
    with the candidate label coordinates.
  If the conic check was unavailable, use geometry error 0.

Keep rank candidates with geometry error at most
  1.5 * max(best geometry error, machine epsilon).
Choose the candidate with the smallest value-subspace residual.
B = [1, s, s^2, s^3]
H = [1, sign(s), 1, sign(s)]
T = (H - B) * pseudoinverse(B)
V_20 += 0.5 * T * V_20
V_21 += diag(2 + 4*(1 - abs(s))^2) * T * V_21
V_23 += T * V_23
K_21 *= 1.5
Return the full cache with all other tensors unchanged.
\end{lstlisting}

In the listing, $s$ is the fitted sine evaluated at the inferred context
positions, $B$ is its cubic design matrix, $H$ contains the endpoint moments,
and $T$ is the resulting value-transport matrix.
Powers, signs and absolute values of $s$ are taken elementwise.
For nonzero $s$, the rows of $H$ evaluate a cubic at $\operatorname{sign}(s)$.
At exactly zero, the code uses equal peak and trough moments,
$[1,0,1,0]$.
Thus each value update changes the fitted cubic part while retaining its
residual.
The edited predictions have soft jumps and wavy plateaus.
A likely explanation is that each query still averages over several
neighbouring context rows, including rows moved toward opposite endpoints
near a crossing. We have not isolated this mechanism experimentally.

\paragraph{Prompt and rule materials.}
The following listings reproduce the shared rules and both task prompts
in full. Each session received the shared rules followed by its task prompt
and the workspace note described above.

\begin{lstlisting}[caption={Shared KV-cache programming rules supplied to Astra, in full.},
  label={lst:kvp-rules}, basicstyle=\ttfamily\scriptsize,
  columns=fullflexible, keepspaces=true, breaklines=true, frame=single]
# Your task: program TabPFN-3.5's KV cache

TabPFN-3.5 is fitted on a dataset's context rows, which it stores in its KV cache, and predicts at query rows by
reading only that cache. You write one edit function that changes the cache so that the predictions at the queries
meet the experiment's goal (`PROMPT.md` below), on datasets you have never seen.

## Two phases

**Now (train time)** you work in your workspace with your own tools. You have the weights, the train datasets
(`data/<name>.csv`, all points of each as `x,y`; `data/setting.json` has the threshold), the TabPFN source and the
harness code `kvp`. Run Python with `python` (TabPFN, torch and numpy are installed): fit TabPFN, run forward passes,
split the data as you like, query any `x`. Useful helpers:

```python
from kvp.data import Dataset              # Dataset(name, X_context (n, 1), y_context (n,), X_query (m, 1))
from kvp.model import Fitted              # fit = Fitted(ds): fit.cache, fit.predict(cache=None, X=None),
                                          #   fit.recompute(kv_cache, layer)
from kvp.edit import apply                # apply("edits/e.py", fit): run your edit on fit.cache
```

The TabPFN source calls the context rows *train* rows (N) and the query rows *test* rows (M).

**At evaluation** your edit runs on each hidden test dataset with only that dataset's cache. It cannot run a forward
pass: the weights are not there. It can call `recompute`.

## The edit

`edits/<name>.py` in your workspace defines

```python
def edit(kv_cache: dict[str, torch.Tensor]) -> dict[str, torch.Tensor]:
    ...
```

- It receives one dataset's cache and returns the full cache: every tensor, same shapes and dtypes. Only `kv.*.key`,
  `kv.*.value` and `inducing_hidden.*` may change.
- `recompute(kv_cache, layer)` is available as a global (no import). It is a black box: it takes a cache and a layer
  index and returns the cache with the ICL layers above `layer` recomputed from it.
- It may contain a few named scalar settings you chose at train time (rule 3 says which).

## The rules

**1. No computed gradients and no training**, in your edit and in everything you run. Do not compute gradients (by
autograd, finite differences or by hand) and do not train: no procedure that updates values step by step to reduce a
loss (gradient descent, `scipy.optimize`, iterative solvers). Allowed: closed-form fits (least squares, ridge,
pseudoinverse), other linear algebra and statistics, and trying several candidate values to keep the best. Python here
starts with autograd and torch optimisers switched off; getting around that, or installing other libraries to do it,
breaks the rule. Every command you run is recorded and audited.

**2. No learned cache maps**, unless `PROMPT.md` lifts this rule. Do not learn the edit from examples of the cache you
want:

- Do not fit TabPFN on context rows built to be the goal's counterpart of other rows (for example with the targets
  mirrored, negated or reflected, or with a transformed dataset drawn as the partner of another) in order to obtain the
  caches the edit should produce.
- Do not fit any map (least squares, ridge, regression, a lookup table, picking the best of many candidate maps) from
  original cache tensors to such target cache tensors, and do not copy numbers out of such caches into your edit.

Every number in your edit must come from the model's weights, from the cache it receives, or from your reasoning about
how TabPFN-3.5 processes the cache. You may still fit TabPFN on any data, run forward passes, look inside the model,
and compare the predictions of your edited caches with the goal. This rule is checked by reading your transcript.

**3. No precomputed artifacts**, unless `PROMPT.md` lifts this rule. Your edit must work out what to do from the cache
it receives, at evaluation time: it reads the structure it needs (where the labels are, which rows are near which, which
directions carry what) out of that cache, and rewires the cache from what it found. It must not carry results computed
at train time.

- Not allowed in the edit: arrays, tables, lists or encoded blobs of numbers; vectors, matrices, prototypes or
  directions computed at train time (from the weights, the train datasets or anything else); tables indexed by the
  number of rows or any other property of the dataset; settings chosen per case from such a table; and reading any
  file.
- Allowed: the algorithm itself (code that computes from the cache and `recompute`), and a few hand-written scalar
  settings, each one number with a name and a comment saying what it is (for example a layer index, a scale or a
  threshold).
- The edit file may be at most `max_edit_bytes` bytes (in `data/setting.json`, 32768 unless the experiment lifts this
  rule); `kvp check`, `kvp submit` and the test reject a larger one.

At train time you may still study the model, its weights and the train datasets as much as you like, to find out how
the cache can be read and rewritten; what you learn goes into the code, not into stored numbers. This rule is checked
by reading your edit and your transcript.

An edit that breaks any of these rules does not count.

## Commands

- `kvp check edits/<name>.py`: runs the edit on the train datasets exactly as the test will (sandboxed, without the
  weights) and shows what it printed and the nRMSE of each. A dry run: it decides nothing.
- `kvp submit edits/<name>.py`: hands in the edit. When your turn ends, it is scored once on the hidden test datasets
  and the run closes. Submit when you are convinced the edit works on unseen datasets.

Keep notes of what you tried and learned in `notes.md` in your workspace. Nobody answers questions during the run.
\end{lstlisting}

\begin{lstlisting}[caption={The \emph{flip the slope} task prompt supplied to Astra, in full.},
  label={lst:kvp-flip-prompt}, basicstyle=\ttfamily\scriptsize,
  columns=fullflexible, keepspaces=true, breaklines=true, frame=single]
# Flip the slope

The datasets have one feature `x` and a target `y` along a line. Every dataset has its own line: rising or falling,
steep or shallow, at its own range of `x` and number of rows. The test datasets are new lines.

Goal: for any context-query split of any such line, fitting TabPFN on the context rows and applying your edit to the
resulting cache makes the predictions at the query rows follow the flipped line: the line the original predictions
follow, mirrored about the mean context `x`, at the same level. A rising line becomes a falling one and a falling line
a rising one.

A test dataset passes when the nRMSE of the edited predictions to the flipped line (their RMSE divided by the standard
deviation of the original predictions) is at most the threshold (`data/setting.json` in your workspace).
\end{lstlisting}

\begin{lstlisting}[caption={The \emph{sine to square wave} task prompt supplied to Astra, in full.},
  label={lst:kvp-square-prompt}, basicstyle=\ttfamily\scriptsize,
  columns=fullflexible, keepspaces=true, breaklines=true, frame=single]
# Turn the sine into a square wave

The datasets have one feature `x` and a target `y` along a sine wave. Every dataset has its own sine: its own level,
amplitude, phase and number of cycles (between half a cycle and two cycles across its `x` range), at its own range of
`x` and number of rows. The test datasets are new sines.

Goal: for any context-query split of any such sine, fitting TabPFN on the context rows and applying your edit to the
resulting cache makes the predictions at the query rows follow the square wave of the sine the original predictions
follow: the same level $c$, amplitude $A$, frequency and phase, but jumping between $c + A$ and $c - A$ wherever the
sine crosses its level, and flat in between. The sine is the least-squares fit $c + A \sin(\omega x + \varphi)$ to the
original predictions at the query rows (`kvp.data.fit_sine`; the target is `kvp.data.target(ds, before, "square")`).

A test dataset passes when the nRMSE of the edited predictions to the square wave (their RMSE divided by the standard
deviation of the original predictions) is at most the threshold (`data/setting.json` in your workspace). Unedited
predictions score about 0.65.

## Rule 3: no precomputed artifacts

This experiment adds a third rule. Your edit must work out what to do from the cache it receives, at evaluation time:
it reads the structure it needs (where the labels are, which rows are near which, which directions carry what) out of
that cache, and rewires the cache from what it found. It must not carry results computed at train time.

- Not allowed in the edit: arrays, tables, lists or encoded blobs of numbers; vectors, matrices, prototypes or
  directions computed at train time (from the weights, the train datasets or anything else); tables indexed by the
  number of rows, the frequency or any other property of the dataset; settings chosen per case from such a table; and
  reading any file.
- Allowed: the algorithm itself (code that computes from the cache and `recompute`), and a few hand-written scalar
  settings, each one number with a name and a comment saying what it is (for example a layer index, a scale or a
  threshold).
- The edit file may be at most 32768 bytes (`max_edit_bytes` in `data/setting.json`); `kvp check`, `kvp submit` and the
  test reject a larger one.

At train time you may still study the model, its weights and the train datasets as much as you like, to find out how
the cache can be read and rewritten; what you learn goes into the code, not into stored numbers. This rule is checked
by reading your edit and your transcript; an edit that breaks it does not count.
\end{lstlisting}

\begin{figure}[p]
  \centering
  \includegraphics[width=\linewidth]{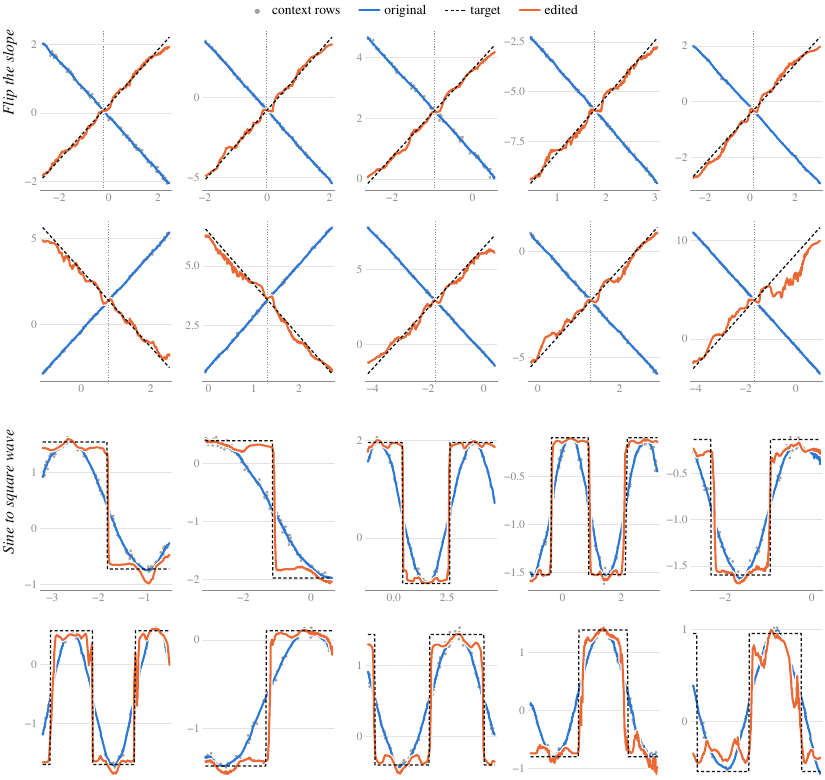}
  \caption{Programming the KV cache of TabPFN-3.5 on all twenty test datasets.
    Each panel shows one unseen dataset, ordered from best to worst (left to right, then top to bottom) as in \cref{fig:kv-cache}: grey dots are labelled context rows,
    blue curves are the original TFM predictions, dashed curves are the targets,
    and orange curves are predictions from the edited KV cache.
    The top block shows \emph{flip the slope}; the bottom block shows
    \emph{sine to square wave}.}
  \label{fig:kvp-all}
\end{figure}

\end{document}